\documentclass{article}

\usepackage{PRIMEarxiv}

\usepackage[utf8]{inputenc} 
\usepackage[T1]{fontenc}    
\usepackage{hyperref}       
\usepackage{url}            
\usepackage{booktabs}       
\usepackage{amsfonts}       
\usepackage{nicefrac}       
\usepackage{microtype}      
\usepackage{lipsum}
\usepackage{fancyhdr}       
\usepackage{graphicx}       
\graphicspath{{media/}}     

\usepackage{amssymb}
\usepackage{amsmath,amsfonts}
\usepackage{algorithmic}
\usepackage{algorithm}
\usepackage{array}
\usepackage{subfigure}
\usepackage{caption}

\usepackage{textcomp}
\usepackage{stfloats}
\usepackage{url}
\usepackage{verbatim}
\usepackage{graphicx}
\usepackage{bbding}
\usepackage{multirow}
\usepackage{makecell} 

\usepackage[square,sort,comma,numbers]{natbib} 

\usepackage{wrapfig} 
\usepackage{flushend}

\usepackage{booktabs} 
\usepackage{pifont}

\usepackage{amsmath} 
\usepackage{amsfonts}

\title{FeTT: Continual Class Incremental Learning via Feature Transformation Tuning} 

\author{
  Sunyuan Qiang \\
  Macau University of Science and Technology \\
  Macau SAR, China \\ \texttt{3220004460@student.must.edu.mo} \\
   \And
  Xuxin Lin \\
  Department of Artificial Intelligence \\ 
  Zhuhai City Polytechnic \\
  Zhuhai, China \\ 
  \texttt{linxuxin6@gmail.com} \\
  \AND 
  Yanyan Liang~\thanks{Corresponding author} \\
  Macau University of Science and Technology \\
  Macau SAR, China \\ 
  \texttt{yyliang@must.edu.mo} \\
  \And
  Jun Wan \\
  Macau University of Science and Technology \\
  MAIS, CASIA and SAI, UCAS \\  
  Macau SAR and Beijing, China \\    
  \texttt{jun.wan@ia.ac.cn} \\
  \And
  Du Zhang \\
  Macau University of Science and Technology \\
  Macau SAR, China \\
  \texttt{duzhang@must.edu.mo} \\
}

\begin{document}
\maketitle

\begin{abstract}
Continual learning (CL) aims to extend deep models from static and enclosed environments to dynamic and complex scenarios, enabling systems to continuously acquire new knowledge of novel categories without forgetting previously learned knowledge. Recent CL models have gradually shifted towards the utilization of pre-trained models (PTMs) with parameter-efficient fine-tuning (PEFT) strategies. However, continual fine-tuning still presents a serious challenge of catastrophic forgetting due to the absence of previous task data. Additionally, the fine-tune-then-frozen mechanism suffers from performance limitations due to feature channels suppression and insufficient training data in the first CL task. To this end, this paper proposes feature transformation tuning (FeTT) model to non-parametrically fine-tune backbone features across all tasks, which not only operates independently of CL training data but also smooths feature channels to prevent excessive suppression. Then, the extended ensemble strategy incorporating different PTMs with FeTT model facilitates further performance improvement. We further elaborate on the discussions of the fine-tune-then-frozen paradigm and the FeTT model from the perspectives of discrepancy in class marginal distributions and feature channels. Extensive experiments on CL benchmarks validate the effectiveness of our proposed method.  
\end{abstract}


\section{Introduction}

Continual learning (CL) empowers deep neural networks (DNNs) to continuously assimilate new knowledge and experiences from evolving data, ensuring adaptability to shifting environments and emerging tasks. Specifically, CL~\cite{DBLP:journals/corr/abs-1904-07734} currently can be further  categorized into three distinct configurations: domain-incremental learning (DIL), task-incremental learning (TIL), and class-incremental learning (CIL). DIL~\cite{DBLP:conf/cvpr/MirzaMPB22} focuses on continual learning from diverse domains within the same category space. TIL~\cite{DBLP:journals/pami/LangeAMPJLST22} emphasizes incremental learning of new classes and supplies task IDs during inference. CIL~\cite{DBLP:journals/corr/abs-2302-03648} directly learns new categories incrementally without task-specific information during inference.

However, the dilemma of catastrophic forgetting~\cite{MCCLOSKEY1989109}, where a model completely forget previously learned knowledge during CL process, significantly limits its applicability. To mitigate this issue, the research community has put forward a variety of CL models, including regularization methods~\cite{DBLP:journals/corr/KirkpatrickPRVD16,DBLP:conf/eccv/AljundiBERT18,DBLP:journals/corr/HintonVD15,DBLP:conf/eccv/LiH16}, memory replay methods~\cite{DBLP:conf/cvpr/RebuffiKSL17,DBLP:conf/nips/Lopez-PazR17,DBLP:conf/aaai/QiangH0L0Z23}, dynamic architecture methods~\cite{DBLP:conf/cvpr/YanX021,DBLP:conf/eccv/WangZYZ22,qiang2024dynamic}, and most recently, pre-trained models (PTMs) based methods~\cite{DBLP:conf/cvpr/0002ZL0SRSPDP22,DBLP:conf/eccv/0002ZESZLRSPDP22,DBLP:conf/cvpr/SmithKGCKAPFK23,DBLP:journals/corr/abs-2303-07338}.

\begin{figure*}[!tb]
\centering   
\includegraphics[width=0.65\textwidth]{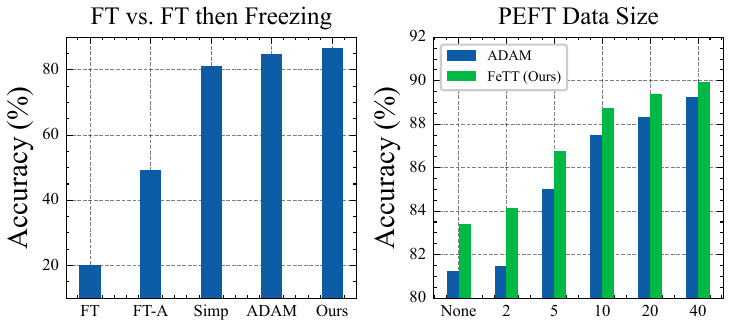}  
\caption{Left: Due to the mismatch between the distributions of training and testing data, continual fine-tuning (FT and FT-Adapter) lead to a serious dilemma of catastrophic forgetting. The fine-tuning and then freezing paradigm (Simp, ADAM, and Ours) demonstrates superior performance. Right: PEFT requires more training data (measured by number of classes) to achieve higher performance. Our method, on the other hand, can enhance its performance without necessitating extra data or training costs.   }  
\label{figures_full_finetune_and_dataset_size} 
\end{figure*}  

Among them, PTMs based CL methods demonstrate strong performance potential due to the excellent feature representation space. However, continual parameter-efficient fine-tuning (PEFT)~\cite{DBLP:conf/cvpr/0002ZL0SRSPDP22} remains ineffective in mitigating the catastrophic forgetting caused by differences in training and test distributions resulting from data absence, as discussed in Section~\ref{section_method_discussion_paradigm}. On the other hand, the fine-tuning in the first task and then freezing the backbone in the following CL tasks based (fine-tune-then-frozen) paradigm~\cite{DBLP:journals/corr/abs-2303-07338}, although it diminishes discrepancies in class marginal data distributions, imposes considerable performance restrictions due to the severely limited PEFT data of the first task. Further discussion in Section~\ref{section_method_discussion_effect_fett} also revealed that fine-tuning on the first task resulted in feature channels overly responsive solely to the data of the first task, thereby suppressing the future activation of other channels. As shown in Figure~\ref{figures_full_finetune_and_dataset_size}, empirical findings highlight a notable disparity in performance between the continual fine-tuning and fine-tune-then-frozen paradigm. Simultaneously, different quantities of PEFT data have a profound impact on the CL performance. The observations above serve as a strong inspiration for proposing a new data-independent tuning method that can be utilized across various tasks in CL scenario. 

To this end, we propose to employ non-parametric transformation functions on backbone features to dynamically adjust feature channel scales, thereby enhancing the performance of all CL tasks. Specifically, the PEFT strategies is initially employed to narrow the gap between the PTMs and downstream datasets in the first task. Subsequently, the channel-wise transformation is performed to smooth the scale of the obtained backbone features. Finally, class prototypes are updated for classification. Our model achieves improvement without the need for additional training data or parameter costs. As shown in Figure~\ref{figures_full_finetune_and_dataset_size}, performance is enhanced under the fine-tune-then-frozen paradigm and different PEFT data sizes. We also conduct extensive experiments covering 6 different datasets and 14 different benchmark settings to validate the outstanding performance of our model. Particularly, we achieve about $93\%$ average accuracy
on CIFAR100 B0 Inc10 (10 tasks) benchmark.   

The main contributions of our work are as follows: 

\begin{itemize} 
    \item We propose employing feature transformation tuning (FeTT) in PTMs based CL scenarios to non-parametrically adjust the backbone features without additional training or parameter overhead. Further extension could involve ensemble strategy utilizing multiple PTMs to enhance performance gains. 
    \item We conduct analysis and discussion of the model, initially finding that the fine-tune-then-frozen paradigm is superior to full PEFT in mitigating class distribution discrepancy, although we also observe the phenomenon of certain feature channels being suppressed, which can be further alleviated by our proposed FeTT method.  
    \item Extensive benchmark experiments and ablation studies across 6 different datasets validate the model's performance. We additionally present comprehensive comparative experimental results demonstrating the integration of FeTT model in a plug-and-play manner with various PEFT strategies.  
\end{itemize}

\section{Related Work}

In this section, we provide a comprehensive overview of existing literature relevant to our proposed method, including continual learning (CL), pre-trained models (PTMs), parameter-efficient fine-tuning (PEFT), and feature transformation.

\subsection{Continual Learning}

Continual learning (CL) is a learning paradigm where knowledge from new coming classes is incrementally added into existing model to adapt the dynamic environments, while preserving previously acquired knowledge from earlier tasks to resist catastrophic forgetting. Currently, there are four main directions~\cite{DBLP:journals/nn/BelouadahPK21,DBLP:journals/pami/LangeAMPJLST22,DBLP:journals/corr/abs-2401-16386} of CL methods: regularization, memory replay, dynamic architecture-based, and pre-trained methods.

Regularization based methods typically impose constraints on both new and old models to alleviate forgetting, such as  constraints on importance weights~\cite{DBLP:journals/corr/KirkpatrickPRVD16,DBLP:conf/eccv/AljundiBERT18} and knowledge distillation (KD)~\cite{DBLP:journals/corr/HintonVD15,DBLP:conf/eccv/LiH16,DBLP:journals/pr/LiWSX23,WU2024110440}. Memory replay methods~\cite{DBLP:conf/cvpr/RebuffiKSL17,DBLP:conf/nips/Lopez-PazR17,SONG2024110506} weaken the strict paradigm of CL mainly by allowing a small subset of old retained task data or synthesized data~\cite{DBLP:conf/nips/ShinLKK17,DBLP:conf/icml/GaoL23a,DBLP:journals/pr/LaoMTDFH21} to be jointly trained with the current task. Further combining the replay with distillation regularization yields more performance improvements~\cite{DBLP:conf/eccv/DouillardCORV20,DBLP:conf/aaai/QiangH0L0Z23}. Dynamic architecture-based methods~\cite{DBLP:conf/cvpr/YanX021,DBLP:conf/eccv/WangZYZ22,qiang2024dynamic} primarily freeze the models trained on previous tasks to mitigate forgetting, while allocating additional parameter space to learn new tasks. Due to the excellent feature representations provided by pre-trained models (PTMs)~\cite{DBLP:conf/iclr/DosovitskiyB0WZ21,DBLP:conf/icml/RadfordKHRGASAM21}, recent CL methods also have gradually benefited from the performance enhancement brought by PTMs, mainly including prompt based CL~\cite{DBLP:conf/cvpr/0002ZL0SRSPDP22,DBLP:conf/eccv/0002ZESZLRSPDP22,DBLP:conf/cvpr/SmithKGCKAPFK23}, zero-shot CL~\cite{DBLP:conf/iccv/ZhengMWQYY23,DBLP:journals/corr/abs-2403-11549}, and prototype CL~\cite{DBLP:journals/corr/abs-2303-07338,DBLP:conf/iccv/ZhangWKCW23,DBLP:conf/nips/McDonnellGPAH23}.

As mentioned above, PTMs based CL methods have shown more promising capabilities. 
However, the prompt based CL strategy~\cite{DBLP:conf/cvpr/0002ZL0SRSPDP22} usually necessitates learnable parameters at each task stage, resulting not only in additional parameter overhead but also posing potential risks of forgetting due to continual training. Zero-shot CL~\cite{DBLP:conf/iccv/ZhengMWQYY23,DBLP:journals/corr/abs-2403-11549} typically incur significant costs in terms of acquiring large text-image training datasets. As for prototype methods, there is a requirement to retain additional random matrices~\cite{DBLP:conf/nips/McDonnellGPAH23} and class covariance matrix~\cite{DBLP:conf/iccv/ZhangWKCW23}. On the contrary, our proposed method directly transforms feature representations without requiring any additional parameters or training costs. Moreover, it can be effectively combined with any parameter-efficient fine-tuning (PEFT) strategies to further enhance performance.

\subsection{Pre-trained Models and Fine-tuning}

The pre-training paradigm of natural language processing (NLP)~\cite{DBLP:conf/naacl/DevlinCLT19,radford2019language} is gradually being transferred to computer vision (CV)~\cite{DBLP:conf/icml/RadfordKHRGASAM21,DBLP:conf/icml/ChenK0H20}, enabling deep models to exhibit exceptional capabilities in image feature representation, such as supervised pre-training~\cite{DBLP:conf/iccv/HeGD19}, self-supervised pre-training~\cite{DBLP:conf/icml/ChenK0H20,DBLP:conf/cvpr/He0WXG20}, and vision-language contrastive pre-training~\cite{DBLP:conf/icml/RadfordKHRGASAM21}. As previously discussed, the outstanding feature representation capability of these pre-trained models (PTMs) has greatly enhanced performance in CL tasks. In this paper, we directly adopt the supervised ImageNet PTM as the feature backbone, based on previous benchmark settings and the alignment between pre-training objectives and downstream classification objectives.

The discrepancy between PTMs and downstream tasks generally necessitates further fine-tuning procedures. Early conventional transfer learning methods involved directly fine-tuning all model parameters~\cite{DBLP:journals/corr/abs-2312-12148}, which inevitably required numerous training iterations and substantial datasets. Then, the introduction of parameter-efficient fine-tuning (PEFT) strategies, such as VPT~\cite{DBLP:conf/eccv/JiaTCCBHL22}, LoRA~\cite{DBLP:conf/iclr/HuSWALWWC22}, SSF~\cite{DBLP:conf/nips/LianZFW22}, and Adapter~\cite{DBLP:conf/nips/ChenGTWSWL22}, not only reduce the expense of fine-tuning training but also adapt PTMs to downstream tasks, leading to broad application prospects. In this paper, following the setup in~\cite{DBLP:journals/corr/abs-2303-07338}, we extensively apply our proposed method in a plug-and-play manner to various PEFT strategies for comprehensive comparison.

\subsection{Feature Transformation}

Feature transformation mainly consists of two categories: parametric transformations that involve learnable parameters, and non-parametric (training-free) transformations. Broadly speaking, for the former, deep models, including MLPs, attention mechanisms~\cite{DBLP:conf/nips/VaswaniSPUJGKP17}, normalization~\cite{DBLP:conf/icml/IoffeS15}, and even the PEFT methods mentioned above~\cite{DBLP:conf/eccv/JiaTCCBHL22,DBLP:conf/nips/LianZFW22}, can all be regarded as transformation operations on features to obtain better latent representations. However, it's evident that achieving a good transformation typically requires training data and an optimization process. On the other hand, training-free transformations directly refine features without requiring any additional parameters or training overhead, mainly including non-linear functions~\cite{DBLP:conf/iclr/YangLX21,DBLP:conf/icml/LuoXX22} and FFT based scale functions~\cite{DBLP:journals/corr/abs-2309-11497}. In this paper, we propose incorporating training-free transformations onto the backbone to enable superior feature space for discrimination in CL scenarios.

\section{Method}

\subsection{Preliminaries}

\subsubsection{Problem Formulation}
\label{section_method_problem}

In continual class-incremental learning paradigm~\cite{DBLP:conf/cvpr/RebuffiKSL17}, 
the training data distribution $P_{\mathcal{D}_t}$ at each incremental task $t$ is composed of $N_t$ sample pairs without any previous memory instances, denoted by $\{(\mathbf{x}_{t,i}, y_{t,i})\} _{i=1}^{N_t}$, where $\mathbf{x}_{t,i}$ and $y_{t,i}$ are the $i$-th sample pair at the $t$-th task from data and target space $\mathcal{X}_t$ and $\mathcal{Y}_t$, respectively. Note that the target spaces at different tasks are assumed to be non-overlapping, i.e., $\mathcal{Y}_i \cap \mathcal{Y}_j = \emptyset$ for $i \not= j$. In CL scenario, our objective is to combat catastrophic forgetting, ensuring that a model $F$ parameterized by $\theta$, once learned on a current task dataset $P_{\mathcal{D}_t}$, retains its classification ability across all previously learned tasks $P_{\mathcal{D}_{1:t}}$.

\subsubsection{Parameter-Efficient Fine-Tuning (PEFT)}
\label{section_method_peft_methods}

In this section, we briefly elaborate on the employed PEFT methods~\cite{DBLP:journals/corr/abs-2303-07338}, including VPT~\cite{DBLP:conf/eccv/JiaTCCBHL22}, SSF~\cite{DBLP:conf/nips/LianZFW22}, and Adapter~\cite{DBLP:conf/nips/ChenGTWSWL22}.

Visual Prompt Tuning (VPT) prepends a small amount of learnable parameters tokens to each transformer blocks's input space. Formally, at $i$-th block layer $L_i$, given patch embeddings $\mathbf{E}_{i-1}$ and the class tokens $\mathbf{x}_{i-1}$, the learnable prompt tokens $\mathbf{P}_{i-1}$ are added for attention interaction, 
\begin{equation}
\begin{aligned}
{\left[\mathbf{x}_i, \underline{\,\,\,\,}, \mathbf{E}_i\right] } & =L_i\left(\left[\mathbf{x}_{i-1}, \mathbf{P}_{i-1}, \mathbf{E}_{i-1}\right]\right). \\
\end{aligned} 
\end{equation} 

Scale \& Shift (SSF) performs the scaling and shifting transformation to modulate the intermediate features. Specifically, given the
input features $\mathbf{x}_\text{in}$, and the scale, shift factors $\gamma$, $\beta$, the output $\mathbf{x}_\text{out}$ is calcuated as, 
\begin{equation} 
    \mathbf{x}_\text{out}=\gamma \otimes \mathbf{x}_\text{in}+\beta .   
\end{equation}

Adapter are typically integrated additively into attention blocks via a bottleneck mechanism, which includes learnable weights for upsampling and downsampling. Assume $\mathbf{x}_\ell$ is the feature at $\ell$-th attention block, we have the following tuning process to obtain the output $\mathbf{x}'_\ell$:  
\begin{equation}
    \mathbf{x}'_\ell = \text{FFN}(\mathbf{x}_\ell) + s \cdot \text{ReLU} \left( \mathbf{x}_\ell \cdot \mathbf{W}_\text{down}  \right) \cdot \mathbf{W}_\text{up} . 
\end{equation}

\subsection{FeTT Model}

\begin{figure*}[!tb] 
\centering 
\includegraphics[width=0.99\textwidth]{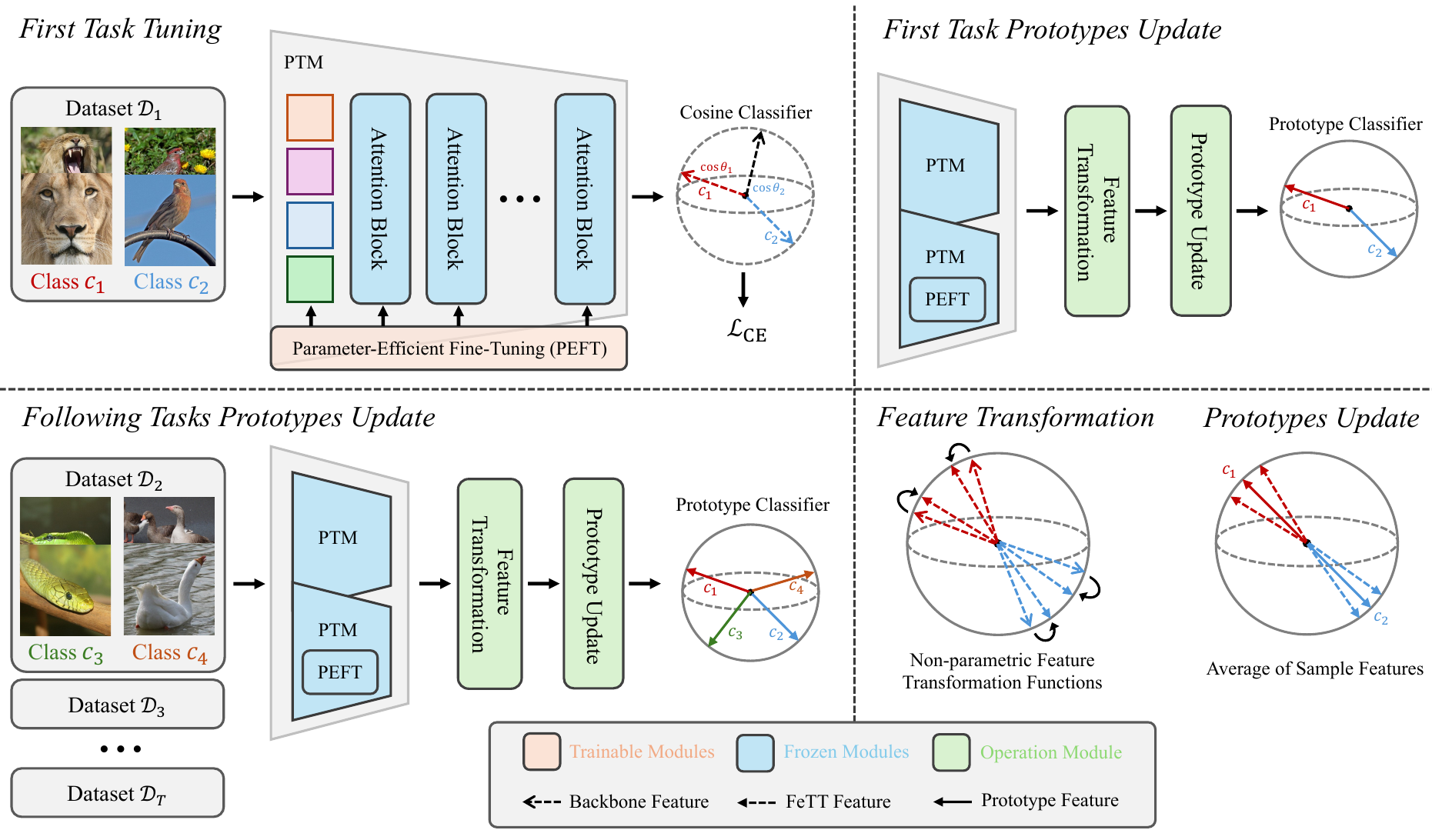} 
\caption{The overall architecture of our proposed model. During the first task, the model applies the PEFT strategies for adapting on downstream tasks. Then, in subsequent incremental tasks (including the first task), the fine-tuned model is frozen and concatenated with the original PTM to update the class prototypes via feature transformations. Feature transformations employ non-parametric functions to modulate backbone features, whereas prototypes are computed by  averaging the sample features within the same category. 
}  
\label{figures_overall_architecture}  
\end{figure*}

The overall architecture is shown in Figure~\ref{figures_overall_architecture}, our model leverages the fine-tune-then-freeze paradigm~\cite{DBLP:journals/corr/abs-2303-07338} of continual learning (CL). In the first task, the training and testing data have consistent class distributions. The fine-tuning process serves to boost the performance of the first task while narrowing the data domain gap for future arriving categories. As for the following incremental tasks, the data distribution discrepancy~\cite{DBLP:conf/aaai/QiangH0L0Z23} of CL and the excellent feature space of PTMs lead us to directly freeze the model, integrating feature transformation and prototype updates for classification prediction. The discussion of the PEFT and discrepancy are elaborated upon in Section~\ref{section_method_discussion}.

Formally, following the notations in Section~\ref{section_method_problem} and fine-tune-then-freeze paradigm~\cite{DBLP:journals/corr/abs-2303-07338}, given data pairs $(\mathbf{x}, y) \sim P_{\mathcal{D}_1}$ in first task and pre-trained model $F$ parameterized by $\theta$, we optimize  
\begin{equation}
    F_{\theta,\phi^*} = \arg \min_\phi  \, \mathbb{E}_{(\mathbf{x},y) \sim P_{\mathcal{D}_1}}  \left[  \ell \left(F_{\theta,\phi} (\mathbf{x}), y \right)\right]  , 
\end{equation}  
where $\phi$ denotes the additional trainable parameters, with PEFT strategies as discussed in Section~\ref{section_method_peft_methods}, aiming to achieve the optimal $\phi^*$ by minimizing the objective loss function $\ell$. The objective loss function is set as cross-entropy, and the classifier metric is cosine similarity.  
\begin{equation}  
    \ell (F_{\theta,\phi}(\mathbf{x}), y) = \ell (\mathbf{z}, y) = - \sum_{i=1}^C y_i \log   \frac{ e^{\cos (\mathbf{z}, \mathbf{w}_y)} }{ \sum_j e^{\cos (\mathbf{z}, \mathbf{w}_j)}} , 
\end{equation}
where $\mathbf{z} = F_{\theta,\phi}(\mathbf{x})$ denotes the backbone feature embeddings. The total number of class is $C$, and $\mathbf{W} = \mathbf{w}_{1:C}$ denotes the classifier weight. Then, the fine-tuned model $F_{\theta,\phi^*}$ is frozen and concatenated with the original PTM $F_{\theta}$ to retain both adaptability to downstream tasks and the original generalization capability~\cite{DBLP:journals/corr/abs-2303-07338}. 
\begin{equation}
    \mathbf{z}' = F'_{\theta,\phi^*} (\mathbf{x} ) =  [ F_{\theta,\phi^*} (\mathbf{x} ),  F_{\theta} (\mathbf{x} )] , 
\end{equation}
where $[,]$ denotes the concatenate operation. 
Next, we introduce two element-wise transformation functions $T(x)$~\cite{tukey1977exploratory,DBLP:conf/iclr/YangLX21,DBLP:conf/icml/LuoXX22} to refine the features $\mathbf{z}'$. 
\begin{equation}
    \text{LogTrans}(x) = T(x) =  \frac{1}{\ln ^\eta \left(\frac{1}{x}+ 1 \right)}  , 
\label{equation_log}
\end{equation} 
\begin{equation}
    \text{PwrTrans}(x) = T(x) =  x^\kappa  , 
\label{equation_pwr}
\end{equation} 
where $\eta$ and $\kappa$ are two hpyer-parameters. Feature transformation tuning (FeTT) are non-parametric in nature, as they modify the channel responses~\cite{DBLP:conf/cvpr/ZhouKLOT16,DBLP:conf/iclr/BaiZJXM021} of features directly to suit both previous and future tasks, without considering the impact of  discrepancy in data distributions, leading to performance improvements. When combined with the PEFT strategy, the fusion of parametric PEFT and non-parametric FeTT results in further significant performance gains. Further detailed discussion can be found in Section~\ref{section_method_discussion_effect_fett}. In practice, these two functions serve similar effects, and we directly employ the former function in the FeTT model, while for ensemble strategies, we apply the two different transformations to the models separately. Then, we get the FeTT features: 
\begin{equation}
    \mathbf{z}'' = \text{LogTrans}(\mathbf{z}'),  \quad \text{or} \quad   \mathbf{z}'' = \text{PwrTrans}(\mathbf{z}') . 
\end{equation}
Finally, in each task, update the class prototype by averaging all FeTT features of the same category as the classifier weights. 
\begin{equation}
    \mathbf{w}_{y} = \frac{1}{N_y} \sum_{i=1}^{N_y} \mathbf{z}''_y ,
\end{equation}
where $\mathbf{z}''_y$ and $N_y$ denote the FeTT features and number of samples in class $y$, respectively. During evaluation, we perform classification using cosine similarity. 
\begin{equation} 
     p_{y} =  \frac{ e^{\cos (\mathbf{z}'', \mathbf{w}_y)} }{ \sum_j e^{\cos (\mathbf{z}'', \mathbf{w}_j)}} . 
\end{equation} 
Additionally, we extend the model to an ensemble strategy for FeTT-E, combining scores from two distinct PTMs for classification.  
\begin{equation} 
     p_{y} =  \frac{ e^{\cos (\mathbf{z}_0'', \mathbf{w}_y) + \cos (\mathbf{z}_1'', \mathbf{w}_y) } }{ \sum_j e^{\cos (\mathbf{z}_0'', \mathbf{w}_j) + \cos (\mathbf{z}_1'', \mathbf{w}_j) }} , 
\end{equation}  
where $\mathbf{z}_0''$ and $\mathbf{z}_1''$ are FeTT features from different two PTMs. Various PTMs exhibit unique inductive biases towards downstream tasks. The introduction of diverse models is intended to broaden the feature space of high-performing PTMs. Note in experimental comparisons, we also ensure fairness by presenting results from the same single PTM for a fair comparison.

\subsection{Model Discussion}
\label{section_method_discussion}

In this section, we present a detailed discussion on the fine-tune-then-freeze paradigm and our FeTT model in the context of continual learning (CL).

\subsubsection{Fine-Tune-Then-Freeze Paradigm}  
\label{section_method_discussion_paradigm}

Following the notations in Section~\ref{section_method_problem}, when conducting supervised training with data, our aim is to find the function within the hypothesis space $F \in \mathcal{F}$ that minimizes the expected risk error $\bar{\mathcal{R}}(F) = \mathbb{E}_{\bar{P}_{\mathcal{D}}} \left[  \ell(F(x),y)\right]$. Then, based on the good approximation of the population distribution $\bar{P}_{\mathcal{D}}$ by the empirical distribution ${P}_\mathcal{D}$, the empirical risk minimization (ERM)~\cite{DBLP:conf/nips/Vapnik91} suggests to minimize the risk on empirical data distribution to approximate the expected risk ${\mathcal{R}}(F) = \mathbb{E}_{(\mathbf{x},y)\sim {P}_\mathcal{D}} \left[  \ell(F(x),y)\right]$. Under the MUD framework~\cite{DBLP:conf/aaai/QiangH0L0Z23} without memory replay samples in CL scenario, the data class marginal distributions of different tasks $t$ degrade to
\begin{equation}
    {P}_{\mathcal{D}_t} (y) = \mathbf{U}(|\mathcal{Y}_{1:t-1}|,|\mathcal{Y}_{1:t}|) , 
\end{equation}  
where $\mathbf{U}$ denotes the uniform distribution. $|\mathcal{Y}_{1:t-1}|$ and $|\mathcal{Y}_{1:t}|$ denote the number of classes in previous tasks and all classes at $t$-th task, respectively. However, the ideal population distribution or test distribution is uniformly distributed across all categories ${P}^\text{test}_{\mathcal{D}_t}  (y) = \mathbf{U}(0,|\mathcal{Y}_{1:t}|)$. The full fine-tuning and PEFT methods mentioned above all based on the ${P}_{\mathcal{D}_t}$, and overlook discrepancy between training distributions, ${P}_{\mathcal{D}_t}$ and ${P}^\text{test}_{\mathcal{D}_t}$. 
\begin{equation}
    \arg \min_\theta  \, \mathbb{E}_{(\mathbf{x},y) \sim {P}_{\mathcal{D}_t} } \left[  \ell \left(F_\theta (\mathbf{x}), y   \right)\right], 
\end{equation}
\begin{equation}
    \arg \min_\phi  \, \mathbb{E}_{(\mathbf{x},y) \sim {P}_{\mathcal{D}_t} } \left[  \ell \left(F_{\theta,\phi} (\mathbf{x}), y   \right)\right] . 
\end{equation} 
The continual training disrupts the excellent representation space of the PTMs, leading to the dilemma of catastrophic forgetting, as illustrated in Figure~\ref{figures_full_finetune_and_dataset_size}.

Additionally, during the first task where the class spaces of the training and testing distributions are consistent (i.e., ${P}_{\mathcal{D}_1} = {P}^\text{test}_{\mathcal{D}_1} = \mathbf{U}(0,|\mathcal{Y}_{1}|)$), it can be considered that the empirical distribution is a good approximation. Minimizing the risk error narrows the gap of downstream tasks and leads to performance improvement. In summary, our model adopts the Fine-Tune-Then-Frozen continual learning paradigm, where PEFT is performed in the first task, followed by model freezing to leverage the excellent feature space of PTMs for prototype updated classification.

\subsubsection{Effect of FeTT}
\label{section_method_discussion_effect_fett}

\begin{figure*}[!tb]
\centering 
\subfigure[Baseline.]{ 
    \begin{minipage}[t]{0.325\textwidth}  
    \centering     \includegraphics[width=1\columnwidth] {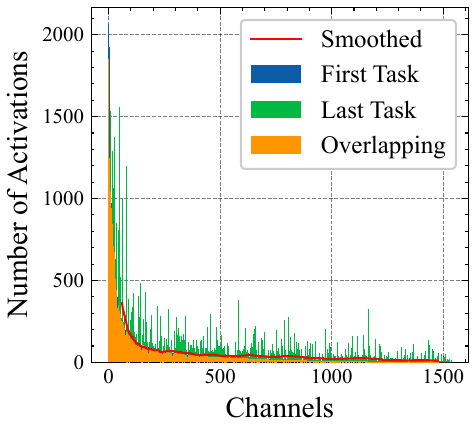}   
    \end{minipage} \label{figure_plots_channel_bar_baseline} 
    }            
    \subfigure[FeTT (Ours). ]{ 
    \begin{minipage}[t]{0.329\textwidth}   
    \centering   \includegraphics[width=1\columnwidth]{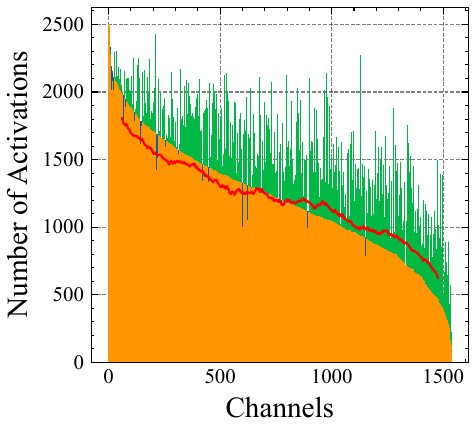}  
    \end{minipage}  \label{figure_plots_channel_bar_fett}  
    }
\caption{The activation frequency of feature embeddings of Adapter based PEFT baseline model and our proposed FeTT model on CIFAR dataset, including fine-tuned feature data from the first task and coming new feature data from the last task. Channels are sorted in a descending order of activation frequency of first task samples. We  additionally include a line plot depicting the moving average of the last task feature activations for better comparative visualization. } 
\label{figure_fett_channel_activations}
\end{figure*}

In this section, we further analyze the effect of FeTT model. Intuitively, different feature channels exhibit distinct responses to different categories~\cite{DBLP:conf/cvpr/ZhouKLOT16}; therefore, in CL scenario, there should naturally be differences in the channel activation patterns between newly arrived categories and the existing categories. Inspired by~\cite{DBLP:conf/iclr/BaiZJXM021}, we extend our analysis to examine the activation frequency patterns of channels in different CL tasks. Specifically, based on the features of the first and last tasks in the CIFAR100 B0 Inc5 benchmark, a channel is deemed activated if its activation value exceeds a certain threshold (e.g., 10\% of the maximum activation value over all  channels). Subsequently, we calculate the activation frequency of channels for different tasks and then sort them  in a descending order of frequency in first task feature samples, as shown in Figure~\ref{figure_fett_channel_activations}.

As shown in Figure~\ref{figure_plots_channel_bar_baseline}, due to the adaptation of the PEFT in the first task, the model exhibits a high response to features relevant to the first task classes, while the activation of most feature channels in the tail is suppressed (The activation frequency is below $100$ after $500$-th channel). Nevertheless, these suppressed feature channels may hold discriminative potential for future category discernment. The moving average frequency trend of the last task remains similar to that of the first task. We believe that a considerable number of discriminative channels specific to the last task category are suppressed, leading the model to primarily utilize highly responsive channels of  first task classes for classification, inevitably limiting the CL performance.

As for our FeTT model, we can firstly observe that the transformation function $T(x)$ in Eqs~\ref{equation_log} and~\ref{equation_pwr} exhibits similar property~\cite{DBLP:conf/icml/LuoXX22}: when $x>0$, then $T'(x)>0$, $T''(x)<0$, and $\lim_{x \rightarrow 0^{+}}T'(x)=+\infty$. A first-order derivative greater than 0 indicates that the relative relationship of the features remains unchanged. Applying the same transformation to all CL tasks does not significantly alter the previous knowledge, thereby preserving the fundamental discriminative nature of the features. Meanwhile, the channel value close to zero are greatly amplified to enhance the response of suppressed features, preparing in advance for newly arriving classes. Additionally, a second derivative less than zero indicates a further reduction in the response differences between feature channels, thereby alleviating the issue of certain feature channels dominating and suppressing others. As shown in Figure~\ref{figure_plots_channel_bar_fett}, our results indicate that the model exhibits activations across a greater number of feature channels on both first task and last task. On the other hand, the channel order is in descending order of the first task frequency, hence the front channels are more relevant to the first task. Compared with the first task activations frequency, the smoothed line illustrates that the channels related to the first task decreased; instead, the activation of subsequent feature channels increased, revealing that the suppression caused by the first task PEFT is alleviated and enhancing performance.

\section{Experimental Results}

This study conducted extensive experiments, specifically including benchmark experiments, ablation studies, and visual explorations. In this section, we will first describe the datasets and experimental setup, followed by a presentation of the experimental results.

\subsection{Setups}

\subsubsection{Datasets}  

According to the CIL benchmarks~\cite{DBLP:conf/eccv/0002ZESZLRSPDP22,DBLP:journals/corr/abs-2303-07338,DBLP:conf/cvpr/0004TLHWCJ020}, this paper conducted experiments across six datasets, comprising  CIFAR100~\cite{krizhevsky2009learning}, CUB200~\cite{wah2011caltech}, ImageNet-A (IN-A)~\cite{DBLP:conf/cvpr/HendrycksZBSS21}, ImageNet-R (IN-R)~\cite{DBLP:conf/iccv/HendrycksBMKWDD21}, ObjectNet (Obj)~\cite{DBLP:conf/nips/BarbuMALWGTK19}, and VTAB~\cite{zhai2019large}. 
 
CIFAR100~\cite{krizhevsky2009learning} is a dataset of $32 \times 32$ RGB images, comprising a total of 100 classes and 60,000 images, where 50,000 are used for training and 10,000 for testing. CUB200~\cite{wah2011caltech} is a dataset for fine-grained bird species visual classification tasks, consisting of a total of 200 categories. ImageNet-R~\cite{DBLP:conf/iccv/HendrycksBMKWDD21} and ImageNet-A~\cite{DBLP:conf/cvpr/HendrycksZBSS21} can be considered as two 200-class datasets evolved from the standard ImageNet~\cite{DBLP:conf/cvpr/DengDSLL009} dataset, where the former incorporates various artistic renditions, while the latter focuses on natural adversarial examples. ObjectNet~\cite{DBLP:conf/nips/BarbuMALWGTK19} consists of a total of 313 categories, with objects in the collected images appearing in cluttered natural scenes and exhibiting unusual poses. In this paper, we follow the settings in~\cite{DBLP:journals/corr/abs-2303-07338} and select subset of 200 classes for CIL scenario. VTAB~\cite{zhai2019large} was originally designed as a cross-domain evaluation benchmark,  comprising a total of 19 datasets divided into three groups: natural, specialized, and structured. Following the setup described in~\cite{DBLP:journals/corr/abs-2303-07338}, we select five datasets from these groups to create a task for cross-domain class-incremental learning.  

\subsubsection{Benchmarks}

Currently, the evaluation benchmark protocols for CIL mainly consist of two aspects~\cite{DBLP:conf/cvpr/RebuffiKSL17,DBLP:conf/cvpr/YanX021}. On one hand, the dataset is partitioned directly into incremental tasks, where all categories are uniformly divided, to facilitate continual class-incremental learning. On the other hand, the model first learns half of the categories (base categories) in the dataset, followed by incremental learning on the remaining half of the categories. The aforementioned protocols are denoted as the Base 0/Half, incremental $n$ task (abbreviated as B 0/H Inc $n$), with $n$ representing the number of classes learned per incremental task. The  Base 0 and Base Half respectively denote conducting incremental tasks directly and learning half of the classes in the first incremental task.

In the example of the CIFAR dataset as shown in Table~\ref{table_comparison_2}, CIFAR B0 Inc10 indicates direct incremental learning of 10 classes per task, resulting in a total of  10 incremental tasks among 100 classes. On the other hand, CIFAR BH Inc10 signifies an initial base task of learning with half of the classes (50 classes), followed by the remaining 50 classes divided into 5 tasks, each involving the learning of 10 incremental classes per task. In total, there are 6 tasks comprising of 1 (initial base task) and 5 (subsequent incremental tasks). Finally, we build a comprehensive and detailed comparison
benchmarks that includes a total of 14 different settings.

\subsubsection{Evaluation Metrics}  

The evaluation metrics used in this paper are all based on the top-1 classification accuracy.  Specifically, we report the average accuracy $\bar{\mathcal{A}} = \frac{1}{T} \sum_{t=1}^T \mathcal{A}_t$ across all tasks as the main quantitative metric for all settings, where $\mathcal{A}_t$ denotes the accuracy on all previously learned tasks after the $t$-th task and $T$ denotes the total number of tasks. Additionally, the last accuracy $\mathcal{A}_T$ after completing  all incremental tasks is also considered for comprehensive evaluation.

\subsection{Implementation Details}

We implement all our model with PyTorch~\cite{DBLP:conf/nips/PaszkeGMLBCKLGA19}. During model parameter-efficient fine-tuning (PEFT) in the first task, we employ the SGD optimizer with a weight decay of 5e-4 across all datasets. The learning rate is set to 0.01 and a cosine annealing schedule is employed. Additionally, the total number of epochs is set to 20, with a batch size of 48. The standard data pre-processing procedures, such as random crop and horizontal flip, are employed to ensure that input images are scaled to $224\times 224$ dimensions. For the pre-trained models (PTMs), we follows the same benchmark settings as before~\cite{DBLP:journals/corr/abs-2303-07338}, using the ImageNet-21k pre-trained Vision Transformer (ViT-B/16) as feature extractor backbone. When employing ensemble strategies,  another model incorporates ImageNet-1k based pre-trained ViT to enhance ensemble diversity. The category order of all incremental tasks remains consistent with prior work~\cite{DBLP:journals/corr/abs-2303-07338}, utilizing a fixed seed for randomization. The hyper-parameters $\eta$ and $\kappa$ of our proposed transformation tuning are mainly set to $0.1$ and $0.3$, respectively.

\subsection{Main Results}

Our proposed method is compared with various other models, including full fine-tuning, LwF~\cite{DBLP:conf/eccv/LiH16}, L2P~\cite{DBLP:conf/cvpr/0002ZL0SRSPDP22}, DualPrompt~\cite{DBLP:conf/eccv/0002ZESZLRSPDP22}, CODA-Prompt~\cite{DBLP:conf/cvpr/SmithKGCKAPFK23}, and ADAM~\cite{DBLP:journals/corr/abs-2303-07338}. In addition, we further conduct detailed comparisons by seamlessly integrating the FeTT model and ensemble FeTT-E model into various parameter-efficient fine-tuning (PEFT) strategies in a plug-and-play manner. The numerical results of the comparative methods are cited from~\cite{DBLP:journals/corr/abs-2303-07338}. As for the ADAM model~\cite{DBLP:journals/corr/abs-2303-07338}, the results are reported based on selecting the best outcome as indicated in the original paper. For the methods marked with $\dag$, the results are based on our re-implementation using open-source code. Note that all the results use the same ImageNet-21k pre-trained ViT-B/16 model, except ours FeTT-E use both ImageNet-21k and ImageNet-1k models for ensemble strategy.

\begin{table*}[!tb]\small 
\centering 
\scalebox{0.8}{  
\begin{tabular}{l|cccccccccccc} 
\toprule 
\multirow{2}{*}{Method} & \multicolumn{2}{c}{CIFAR B0 Inc5} & \multicolumn{2}{c}{CUB B0 Inc10} & \multicolumn{2}{c}{IN-R B0 Inc5} & \multicolumn{2}{c}{IN-A B0 Inc10} & \multicolumn{2}{c}{Obj B0 Inc10} & \multicolumn{2}{c}{VTAB B0 Inc10} \\ 
 &  $\bar{\mathcal{A}}$ & $\mathcal{A}_T$ & $\bar{\mathcal{A}}$ & $\mathcal{A}_T$ & $\bar{\mathcal{A}}$ & $\mathcal{A}_T$ & $\bar{\mathcal{A}}$ & $\mathcal{A}_T$ & $\bar{\mathcal{A}}$ & $\mathcal{A}_T$ & $\bar{\mathcal{A}}$ & $\mathcal{A}_T$ \\
\midrule 
Fine-Tune & $38.90$ & $20.17$ & $26.08$ & $13.96$ & $21.61$ & $10.79$ & $21.60$ & $10.96$ & $19.14$ & $8.73$ & $34.95$ & $21.25$ \\
Fine-Tune Adapter & $60.51$ & $49.32$ & $66.84$ & $52.99$ & $47.59$ & $40.28$ & $43.05$ & $37.66$ & $50.22$ & $35.95$ & $48.91$ & $45.12$ \\
LwF~\cite{DBLP:conf/eccv/LiH16} & \, $46.29$ \, & \, $41.07$ \, & \, $48.97$ \, & \, $32.03$ \, & \, $39.93$ \, & \, $26.47$ \, & \, $35.39$ \, & \, $23.83$ \, & \, $33.01$ \, &  \, $20.65$ \, & \, $40.48$ \, & \, $27.54$ \,  \\  
L2P~\cite{DBLP:conf/cvpr/0002ZL0SRSPDP22} & $85.94$ & $79.93$ & $67.05$ & $56.25$ & $66.53$ & $59.22$ & $47.16$ & $38.48$ & $63.78$ & $52.19$ & $77.11$ & $77.10$ \\
DualPrompt~\cite{DBLP:conf/eccv/0002ZESZLRSPDP22} & $87.87$ & $81.15$ & $77.47$ & $66.54$ & $63.31$ & $55.22$ & $52.56$ & $42.68$ & $59.27$ & $49.33$ & $83.36$ & $81.23$ \\
CODA-Prompt~\cite{DBLP:conf/cvpr/SmithKGCKAPFK23} & $89.11$ & $81.96$ & $84.00$ & $73.37$ & $64.42$ & $55.08$ & $48.51$ & $36.47$ & $66.07$ & $53.29$ & $83.90$ & $83.02$ \\
ADAM$^*$~\cite{DBLP:journals/corr/abs-2303-07338} & $90.65$ & $85.15$ & $92.21$ & $86.73$ & $72.35$ & $64.33$ & $62.81$ & $51.48$ & $69.15$ & $56.64$ & $87.47$ & $85.36$ \\
\midrule 
SimpleCIL$^\dag$~\cite{DBLP:journals/corr/abs-2303-07338} & $87.57$ & $81.26$ & $92.23$ & $86.77$ & $62.39$ & $54.33$ & $60.63$ & $48.45$ & $65.45$ & $53.59$ & $86.34$ & $84.46$ \\ 
+ FeTT (Ours) & $\mathbf{89.22}$ & $\mathbf{83.42}$  & $\mathbf{92.41}$ & $\mathbf{87.02}$ & $64.32$ & $56.55$ & $63.58$ & $52.34$ & $67.13$ & $54.68$ & $88.83$ & $87.61$ \\ 
+ FeTT-E (Ours) & $88.81$ & $83.12$ & $92.08$ & $86.94$ & $\mathbf{68.66}$ & $\mathbf{61.97}$ & $\mathbf{65.93}$ & $\mathbf{54.71}$ & $\mathbf{67.20}$ & $\mathbf{55.05}$ & $\mathbf{88.97}$ & $\mathbf{87.82}$ \\ 
\midrule  
ADAM (VPT)$^\dag$~\cite{DBLP:journals/corr/abs-2303-07338} & $85.28$ & $78.24$ & $91.79$ & $85.92$ & $55.75$ & $47.08$ & $48.62$ & $37.39$ & $62.03$ & $49.17$ & $82.18$ & $79.94$ \\ 
+ FeTT (Ours) & $88.00$ & $82.40$ & $\mathbf{91.80}$ & $85.96$ & $65.79$ & $57.62$ & $\mathbf{57.85}$ & $\mathbf{46.15}$ & $67.24$ & $54.60$ & $85.86$ & $83.65$  \\ 
+ FeTT-E (Ours) & $\mathbf{88.58}$ & $\mathbf{82.84}$ & $91.73$ & $\mathbf{86.22}$ & $\mathbf{68.95}$ & $\mathbf{61.12}$ & $55.37$ & $44.04$ & $\mathbf{68.22}$ & $\mathbf{55.76}$ & $\mathbf{86.63}$ & $\mathbf{85.10}$  \\ 
\midrule  
ADAM (SSF)$^\dag$~\cite{DBLP:journals/corr/abs-2303-07338} & 
$87.89$ & $81.88$ & $91.80$ & $86.43$ & $68.99$ & $60.63$ & $61.04$ & $49.11$ & $69.12$ & $56.38$ & $86.31$ & $82.55$ \\ 
+ FeTT (Ours) & $\mathbf{89.05}$ & $83.67$ & $\mathbf{91.81}$ &$86.43$ & $73.65$ & $66.10$ & $64.03$ & $52.47$ & $\mathbf{70.14}$ & $\mathbf{57.30}$ & $89.50$  & $87.64$ \\  
+ FeTT-E (Ours) & $88.94$ & $\mathbf{83.74}$ & $91.69$ & $\mathbf{86.64}$ & $\mathbf{75.32}$ & $\mathbf{68.00}$ & $\mathbf{67.31}$ & $\mathbf{55.04}$ & $69.91$ & $57.09$ & $\mathbf{90.27}$ & $\mathbf{88.46}$  \\ 
\midrule 
ADAM (Adapter)$^\dag$~\cite{DBLP:journals/corr/abs-2303-07338} & $90.58$ & $85.04$ & $92.24$ & $86.73$ & $62.85$ & $54.78$ & $60.78$ & $48.65$ & $67.15$ & $55.21$ & $86.24$ & $84.38$  \\    
+ FeTT (Ours) & $91.78$ & $86.76$ & $\mathbf{92.32}$ & $\mathbf{86.94}$ & $64.78$ & $57.15$ & $63.53$ & $52.21$ & $69.05$ & $56.83$ & $88.82$ & $87.63$  \\ 
+ FeTT-E (Ours) & $\mathbf{91.96}$ & $\mathbf{86.94}$ & $92.07$ & $\mathbf{86.94}$ & $\mathbf{69.04}$ & $\mathbf{62.38}$ & $\mathbf{65.90}$ & $\mathbf{54.64}$ & $\mathbf{69.46}$ & $\mathbf{57.26}$ & $\mathbf{88.97}$ & $\mathbf{87.82}$ \\ 
\bottomrule 
\end{tabular} 
} 
\caption{Comparison results of average  accuracy $\bar{\mathcal{A}}$ and last accuracy $\mathcal{A}_T$ on six datasets. Note that IN-R, IN-A, and Obj denote the abbreviations of ImageNet-R, ImageNet-A, and ObjectNet datasets, respectively. $*$ mark indicates the best results cited from the original paper. $\dag$ mark denotes the re-implemented results using open source code.}
\label{table_comparison_1}
\end{table*}
\begin{figure*}[!tb]
\centering    
\includegraphics[width=0.99\textwidth]{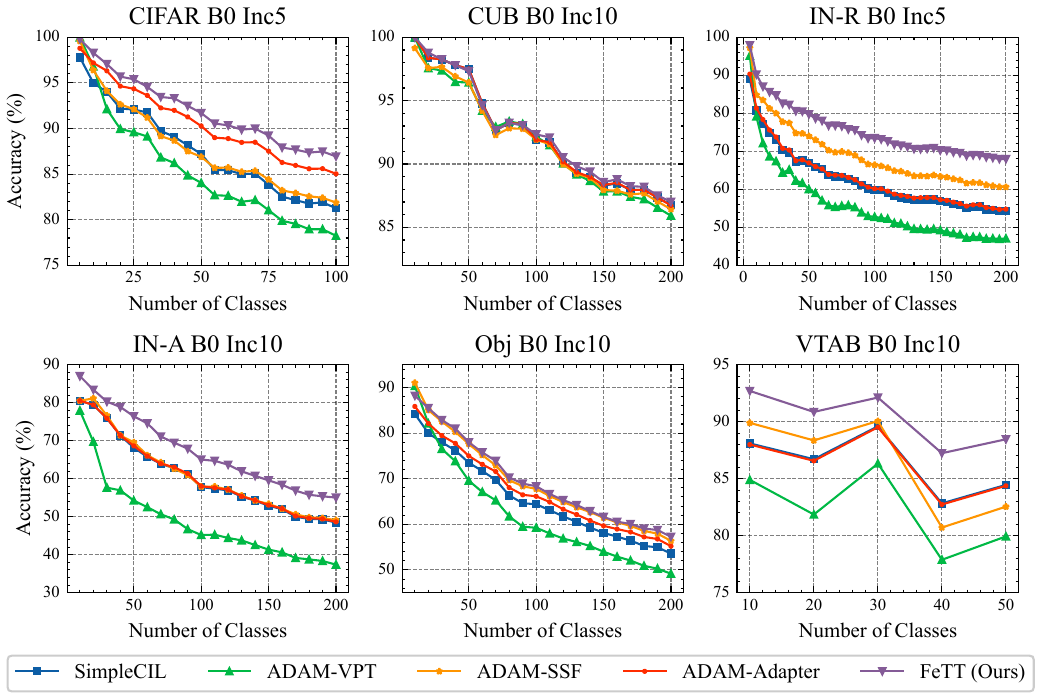}  
\caption{Performance comparison of each step. Our proposed FeTT model directly select the best results for comparison among various parameter-efficient fine-tuning (PEFT) strategies. }   
\label{figures_each_steps1} 
\end{figure*}

Table~\ref{table_comparison_1} and Figure~\ref{figures_each_steps1} summarize the main quantitative results of various methods across different datasets. Overall, with the support of our proposed model, 
whether employing the SimpleCIL method of classifying prototypes using frozen PTMs, or different PEFT-based ADAM strategies, we can observe a noticeable improvement in both average and last accuracy performance, validating the effectiveness of our proposed model. Furthermore, the accuracy line plots for each task in Figure~\ref{figures_each_steps1} demonstrate that our method consistently outperforms other methods, indicating strong performance improvements across all task stages. Particularly, on CIFAR B0 Inc5 benchmark, employing our FeTT method with the SimpleCIL by prototypes classification leads to the enhancement from $87.57 \%$ to $89.22 \%$ ($+1.65 \%$) and $81.26 \%$ to $83.42 \%$ ($+2.16 \%$) on average and last accuracy, respectively. 
With the adapter PEFT strategy, our FeTT-E model demonstrate the improvement in average and last accuracy from $90.58 \%$ to $91.96 \%$ ($+1.38 \%$) and $85.04 \%$ to $86.94 \%$ ($1.90 \%$). Additionally, on cross-domain CIL VTAB benchmark, we achieve performance with average and last accuracy of
$90.27 \%$ and $88.46 \%$ based on SSF PEFT strategy. In general, firstly, by directly using prototypes classification and freezing the model parameters in SimpleCIL, our method already  achieves performance gains through feature transformation. Then, after the PEFT adapter strategy in the first task, the performance of the model is initial improved owing to the mitigation of data domain gap. Moreover, with  our method, the performance is once again enhanced, confirming the generalizability for plug-and-play use in different scenarios.

\begin{table*}[!tb]\small 
\centering
\scalebox{1.0}{  
\begin{tabular}{l|cccccccccc} 
\toprule
\multirow{2}{*}{Method} & \multicolumn{2}{c}{CIFAR} &  \multicolumn{2}{c}{IN-R} & \multicolumn{2}{c}{IN-A} & \multicolumn{2}{c}{Obj}  \\
 & B0 Inc10 & BH Inc10 & B0 Inc10 & BH Inc10  & B0 Inc5 & BH Inc10 & B0 Inc20 & BH Inc20 \\ 
\midrule
SimpleCIL$^\dag$~\cite{DBLP:journals/corr/abs-2303-07338} & $87.13$ & $83.87$ & $61.82$ & $56.69$ & $61.20$ & $53.05$ & $64.58$ & $58.59$ \\
+ FeTT (Ours) & $\mathbf{88.78}$ & $\mathbf{85.82}$ & $63.83$ & $58.69$ & $64.12$ & $57.02$ & $66.23$ & $59.72$ \\
+ FeTT-E (Ours) & $88.40$ & $85.35$ & $\mathbf{68.24}$ & $\mathbf{63.89}$  & $\mathbf{66.62}$ & $\mathbf{58.94}$ & $\mathbf{66.34}$ & $\mathbf{60.09}$  \\    
\midrule
ADAM (SSF)$^\dag$~\cite{DBLP:journals/corr/abs-2303-07338} & $90.72$ & $89.76$ & $73.06$ & $76.04$ & $62.01$ & $64.40$ & $69.95$ & $67.80$ \\ 
+ FeTT (Ours) & $\mathbf{91.51}$ & $\mathbf{90.63}$ & $75.81$ & $79.25$ & $64.47$ & $67.20$ & $\mathbf{70.87}$ & $\mathbf{68.47}$ \\ 
+ FeTT-E (Ours) & $91.48$ & $90.56$ & $\mathbf{77.16}$ & $\mathbf{80.17}$ & $\mathbf{68.23}$ & $\mathbf{69.63}$ & $70.43$ & $68.37$ \\ 
\midrule 
ADAM (Adapter)$^\dag$~\cite{DBLP:journals/corr/abs-2303-07338} & $92.19$ & $91.97$ & $65.57$ & $75.65$ & $61.25$ & $61.73$ & $68.78$ & $68.34$ \\ 
+ FeTT (Ours) & $93.21$ & $92.45$ & $66.70$ & $78.32$ & $63.94$ & $65.28$ & $70.54$ & $68.89$  \\
+ FeTT-E (Ours) & $\mathbf{93.23}$ & $\mathbf{92.47}$ & $\mathbf{71.03}$ & $\mathbf{79.87}$ & $\mathbf{66.51}$ & $\mathbf{68.17}$ & $\mathbf{70.89}$ & $\mathbf{68.87}$ \\ 
\bottomrule 
\end{tabular}
} 
\caption{Comparison results of average  accuracy $\bar{\mathcal{A}}$ on four datasets with more different evaluation settings. Similarly, IN-R, IN-A, and Obj denote the abbreviations of ImageNet-R, ImageNet-A, and ObjectNet datasets, respectively. $\dag$ mark denotes the re-implemented results using open source code.} 
\label{table_comparison_2}
\end{table*} 

\begin{figure*}[!tb]
\centering     
\includegraphics[width=0.99\textwidth]{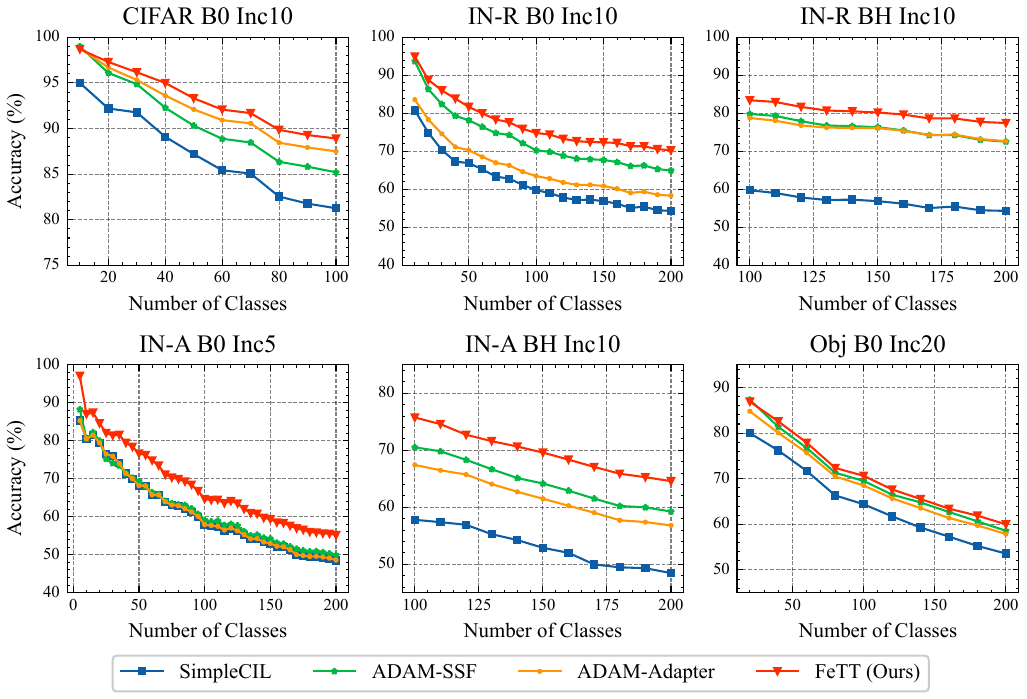} 
\caption{Performance comparison of each step. Our proposed FeTT model directly select the best results for comparison among various parameter-efficient fine-tuning (PEFT) strategies. }   
\label{figures_each_steps2}  
\end{figure*}

As shown in Table~\ref{table_comparison_1}, it is noteworthy that our FeTT model can bring performance improvements over all baseline models. However, when using the ensemble  strategy with the FeTT-E model, it primarily demonstrates further performance on the ImageNet-R, ImageNet-A, and VTAB datasets compared to FeTT alone, yet it does not consistently yield performance gains across all datasets like CIFAR and CUB. We believe that this could be attributed to model preferences in PTMs from ImageNet-1k and ImageNet-21k, where performance differences in downstream tasks may arise due to the gap between downstream datasets and pre-training data. For larger domain gaps and fine-grained image datasets like CIFAR and CUB datasets, the ImageNet-21k based PTMs may be more suitable. Conversely, for downstream tasks derived from standard ImageNet dataset such as ImageNet-A and ImageNet-R, the ImageNet-1k based PTMs might exhibit superior performance. More detailed information about the ablation experiments on PTMs can be found in Tables~\ref{table_ablation_main},~\ref{table_ablation_different_ptms} and~\ref{table_ablation_different_ptms_fett_and_fett_e}.

In addition, Table~\ref{table_comparison_2} and Figure~\ref{figures_each_steps2} summarize further comparative results across various benchmark settings. Similarly, we observe that the performance is further enhanced with the aid of our proposed FeTT model. In detail, on the classic CIFAR B0 Inc10 benchmark with total 10 tasks, we achieve an average accuracy of $93.23 \%$, which validates the success of our method. On another note, in the CIFAR BH Inc10 benchmark setting, it is clearly observed that the PEFT baseline model show a more significant improvement compared to SimpleCIL, possibly due to sufficient data from half of the classes during PEFT in the first task. Despite this, we continue to witness additional performance gains using the FeTT model, advancing from $91.97 \%$ to $92.47 \%$ ($+0.50 \%$), thus confirming the generalization of our proposed feature transformation method. Further experimental results on other datasets, such as ImageNet-A, ImageNet-R, and ObjectNet, in Table~\ref{table_comparison_2} and the line graph of accuracy at each stage in Figure~\ref{figures_each_steps2} also demonstrate the consistent improvement of our method across different datasets.

\subsection{Ablation Study} 

In this section, we conduct comprehensive ablation studies to validate the effectiveness of our proposed method, which mainly included ablation experiments on each component and hyper-parameters.

As shown in Table~\ref{table_ablation_main}, the ablation experiments are performed on the model, respectively, in the baseline model of SimpleCIL with frozen all model parameters, and in the baseline model with the PEFT adapter strategy, to comprehensively validate the model's capabilities. The overall benchmark datasets for model ablation includes the ObjectNet and ImageNet-A datasets. Our proposed method mainly consists of two feature transformation functions and an ensemble strategy, resulting in a total of three components.

Clearly, under the baselines of SimpleCIL and Adapter, we can observe performance improvements brought about by applying two transformation functions. Specifically, the Log feature transformation function yields average accuracies of $67.13 \%$ and $63.58 \%$ on the ObjectNet and ImageNet-A datasets, while the Pwr function achieves average accuracies of $66.99 \%$ and $63.54 \%$. Compared to the baseline model's average accuracies of $65.45 \%$ and $60.63 \%$, both functions show notable performance enhancements, affirming the efficacy of the feature transformation module. When using adapter baseline on the ObjectNet dataset, we can observe that, compared to the baseline model using SimpleCIL, the model's domain gap narrowed via PEFT, leading to a performance improvement from $65.45 \%$ to $67.15 \%$. Furthermore, our proposed two feature transformation functions are able to continue to contribute to performance enhancement. Meanwhile, without employing any feature transformation functions, we introduce diverse pre-trained ImageNet-1k ViT model for ensemble, leading to accuracy boosting compared with two baseline models. Finally, with the support of all constituent components, the model performance achieves the most outstanding results.

\begin{table*}[!tb]\small  
\centering
\begin{tabular}{c|ccc|cccc} 
\toprule
\multirow{2}{*}{Baseline} & \multicolumn{3}{c|}{Ablations} & \multicolumn{2}{c}{Obj} & \multicolumn{2}{c}{IN-A} \\ 
 & Log & Pwr & Ens. & $\bar{\mathcal{A}}$ & $\mathcal{A}_T$ & $\bar{\mathcal{A}}$ & $\mathcal{A}_T$ \\ 
\midrule  
\multirow{5}{*}{SimpleCIL} & & & & $65.45$ & $53.59$ & $60.63$ & $48.45$ \\ 
 & \ding{51} & & & $67.13$ & $54.68$ & $63.58$ & $52.34$ \\
 & & \ding{51} & & $66.99$ & $54.66$ & $63.54$ & $52.34$ \\
 & & & \ding{51} & $65.69$ & $53.74$ & $61.79$ & $51.15$  \\ 
 & \ding{51} & \ding{51} & \ding{51} & $\mathbf{67.21}$ & $\mathbf{55.05}$  & $\mathbf{65.93}$ & $\mathbf{54.71}$ \\ 
\midrule 
\multirow{5}{*}{ADAM (Adapter)} & & & & $67.15$ & $55.21$ & $60.78$ & $48.65$ \\ 
 & \ding{51} & & & $69.05$ & $56.83$ & $63.53$ & $52.21$ \\    
 & & \ding{51} & & $68.96$ & $56.76$ & $63.52$ & $52.21$  \\   
 & & & \ding{51} & $68.00$ & $56.26$ & $61.74$ &  $51.02$\\ 
 & \ding{51} & \ding{51} & \ding{51} & $\mathbf{69.46}$ & $\mathbf{57.26}$ & $\mathbf{65.90}$ & $\mathbf{54.64}$ \\ 
\bottomrule
\end{tabular}
\caption{Ablation results of average  accuracy $\bar{\mathcal{A}}$ and last accuracy $\mathcal{A}_T$ on ObjectNet and ImageNet-A datasets. Log and Pwr are the Log transformation and Power transformation, respectively. Ens. denotes the abbreviations of ensemble strategy.  }
\label{table_ablation_main}
\end{table*}

\begin{figure*}[!tb]   
\centering      
\includegraphics[width=0.75\textwidth]{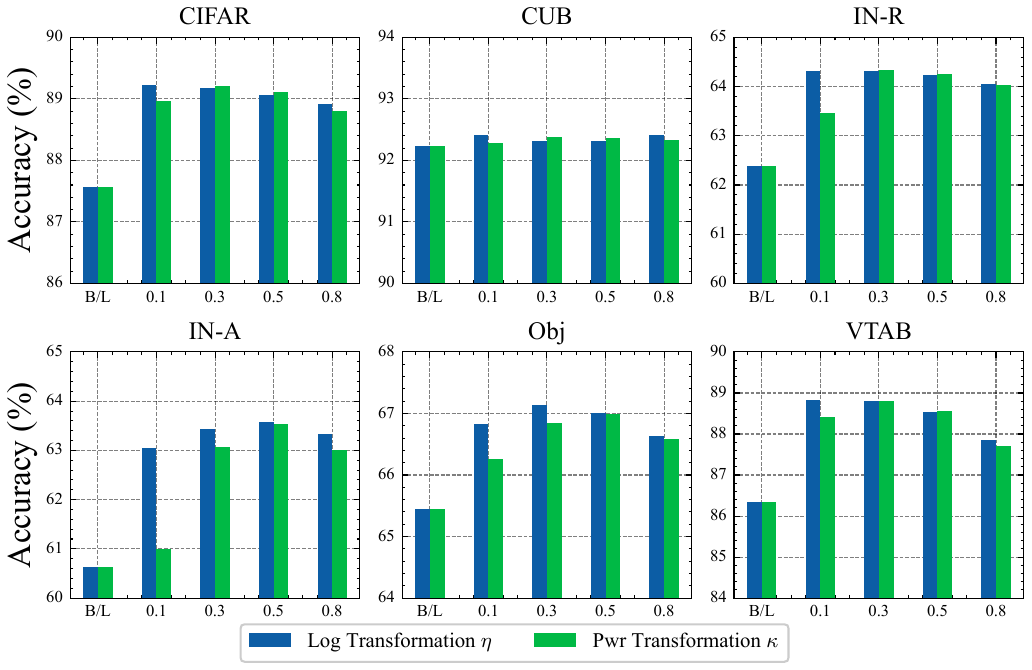}  
\caption{Ablation results of average accuracy $\bar{\mathcal{A}}$ under various hyper-parameters $\eta$ and $\kappa$ varying settings within the interval [0.1, 0.3, 0.5, 0.8]. Here, B/L denotes the baseline model. }   
\label{figures_hyper_params_log_power}  
\end{figure*}

Regarding our two feature transformation functions, in addition to assessing its effectiveness as demonstrated in Table~\ref{table_ablation_main}, it also involves hyper-parameters, namely $\eta$ and $\kappa$, for further analysis and sensitive study. As shown in Figure~\ref{figures_hyper_params_log_power}, to evaluate the hyper-parameters impact, we conduct a systematic investigation by exploring a range of [0.1, 0.3, 0.5, 0.8] across different datasets using SimpleCIL baseline model. In summary, from the bar chart, it can be observed that our transformation function exhibits varying degrees of performance gains across different hyper-parameter settings and datasets. Roughly speaking, with $\eta=0.1$ and $\kappa=0.3$, outstanding performance is demonstrated, highlighting the generalization advantage of our proposed method in hyper-parameter selection. In practice, we mainly set the hyper-parameters $\eta$, $\kappa$ to $0.5$, $0.5$ and $0.3$, $0.5$ on the ImageNet-A and ObjectNet datasets, respectively. Subsequently, in a similar manner, these hyper-parameters results obtained from above SimpleCIL baseline model are directly transferred and applied to other PEFT based models.

\subsection{Further Exploration}

In this section, we conduct an in-depth investigation of our model, including results from different PTMs, analysis of different PEFT dataset sizes, and t-SNE visualization results.

\begin{table*}[!tb]\small
\centering
\begin{tabular}{l|ccccc} 
\toprule
\multirow{2}{*}{Ablations} & \multicolumn{5}{c}{PTMs} \\
 & IN-1K & IN-21K & IN-1K-M & IN-21K-M & CLIP  \\ 
\midrule 
SimpleCIL & $\,$$\,$ $82.79$ $\,$$\,$ & $\,$$\,$ $87.57$ $\,$$\,$  & $\,$$\,$  $86.35$ $\,$$\,$  & $\,$$\,$  $89.75$ $\,$$\,$  & $\,$$\,$  $64.63$ $\,$$\,$   \\
+ Log Trans Func & $\mathbf{85.61}$ & $\mathbf{89.22}$ & $\mathbf{86.48}$ & $\mathbf{89.93}$ & $\mathbf{64.66}$  \\   
\bottomrule
\end{tabular}
\caption{Ablation results of average accuracy $\bar{\mathcal{A}}$ using different pre-trained models (PTMs) on CIFAR B0 Inc5 benchmark. IN-1K and IN-21K denote the ImageNet-1K and ImageNet-21K pre-trained ViTs. IN-1K-M and IN-21K-M denotes the MIIL preprocess based pre-trained ViTs~\cite{DBLP:conf/nips/RidnikBNZ21}. CLIP denotes the pre-trained vision-language model~\cite{DBLP:conf/icml/RadfordKHRGASAM21}.  } 
\label{table_ablation_different_ptms}
\end{table*}

\begin{table*}[!tb]\small 
\centering
\begin{tabular}{l|cccc} 
\toprule
\multirow{2}{*}{Ablations} & \multicolumn{2}{c}{Obj} & \multicolumn{2}{c}{IN-A}  \\  
 & IN-1K & IN-21K & IN-1K & IN-21K \\ 
\midrule
SimpleCIL & $63.12$ & $65.45$ & $60.04$ & $60.63$ \\
+ FeTT (Ours) & $65.25$ & $67.13$ & $65.68$ & $63.57$ \\
+ FeTT-E (Ours) & $\mathbf{67.21}$ & $\mathbf{67.21}$ & $\mathbf{65.93}$ & $\mathbf{65.93}$ \\
\bottomrule
\end{tabular}
\caption{Ablation results of average accuracy $\bar{\mathcal{A}}$ using ImageNet-1K and ImageNet-21K pre-trained ViT models on ObjectNet and ImageNet-A benchmarks. }
\label{table_ablation_different_ptms_fett_and_fett_e} 
\end{table*}

As shown in Table~\ref{table_ablation_different_ptms}, we explore the application of our feature transformation function across various types of PTMs, including ImageNet based ViT-B/16 PTMs, ImageNet-MIIL~\cite{DBLP:conf/nips/RidnikBNZ21} based ViT-B/16 PTMs, as well as CLIP~\cite{DBLP:conf/icml/RadfordKHRGASAM21} based ViT-B/32 PTMs. Obviously, thanks to the training data refinement, the ImageNet-MIIL based PTMs achieve better performance. On the contrary, the performance of the CLIP model is likely weakest due to the significant gap between the text-image training dataset and the CIFAR dataset. In this paper, we mainly adopt the ImageNet-21k based PTMs for fair comparison, although stronger models generally have the potential to achieve better performance in CIL tasks. More importantly, applying our feature transformation function to various PTMs consistently shows performance gains in Table~\ref{table_ablation_different_ptms},  underscoring its broad applicability and generalizability. Meanwhile, Table~\ref{table_ablation_different_ptms_fett_and_fett_e} further illustrates the experimental results of single models and ensemble models on the ObjectNet and ImageNet-A datasets using IN-1k and IN-21k PTMs. The performance improvement of FeTT-E model over individual models suggests the promising potential of integrating diverse models.

\begin{table*}[!tb]\small  
\centering 
\begin{tabular}{l|cccccc} 
\toprule 
\multirow{2}{*}{Ablations} & \multicolumn{6}{c}{Number of training classes in first step for PEFT} \\ 
 & None & 2 classes & 5 classes & 10 classes & 20  classes & 40  classes  \\  
\midrule 
ADAM (Adapter) & $\,$ $81.26$ $\,$ & $81.48$ & $85.03$ & $87.50$ & $88.33$ & $89.27$  \\ 
+ FeTT (Ours) & $\mathbf{83.42}$ & $\mathbf{84.16}$ &  $86.75$ & $88.76$ & $89.39$ & $89.93$ \\ 
+ FeTT-E (Ours) & $83.12$ & $83.79$ & $\mathbf{86.96}$ & $\mathbf{88.91}$ & $\mathbf{89.61}$ & $\mathbf{90.17}$ \\ 
\bottomrule  
\end{tabular}
\caption{Ablation results of last accuracy $\mathcal{A}_T$ using different parameter-efficient fine-tuning (PEFT) data volume in the first step (measured by number of classes) on the CIFAR dataset. The None signifies the absence of any training data, which degrades to the SimpleCIL method. } 
\label{table_different_data_volume}
\end{table*}

As for the results of dataset sizes, Table~\ref{table_different_data_volume} illustrates the impact of different PEFT data volumes in the first task, which are measured by the number of classes. As the volume of data increases gradually, there is a corresponding steady improvement in last accuracy, potentially indicates the performance gains brought about by the PEFT strategy of Adapter. By employing our proposed FeTT model, we can continue to observe further performance improvements,  revealing that our non-learnable feature transformation functions demonstrate performance decoupling from increases in data volume of Adapter, thus proving the effectiveness and generalization. 

\begin{figure*}[!tb]  
\begin{center} 
\subfigure[t-SNE. Baseline.]{ 
    \begin{minipage}[t]{0.325\textwidth}  
    \centering     \includegraphics[width=1\columnwidth]  {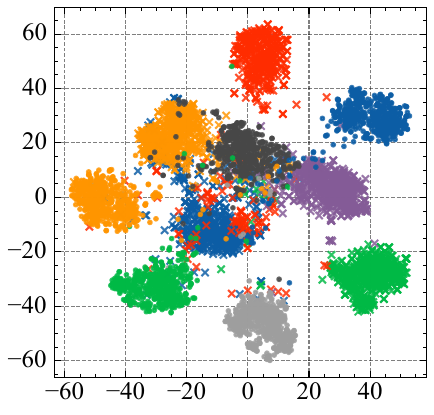}   
    \end{minipage} \label{figure_plots_tsne_baseline} 
    }            
    \subfigure[t-SNE. FeTT (Ours). ]{
    \begin{minipage}[t]{0.325\textwidth}    
    \centering   \includegraphics[width=1\columnwidth]{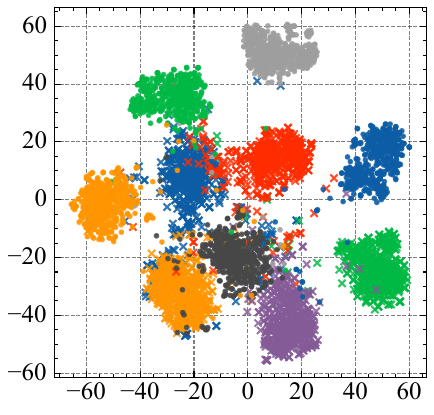} 
    \end{minipage}  \label{figure_plots_tsne_ours} 
    } 
\end{center} 
\caption{The t-SNE visualization results on the CIFAR B0 Inc5 benchmark. The distinct marker symbols and colors are utilized to symbolize different class samples. The five classes in the first base step are denoted with circles $\circ$, while the other five classes are in the last incremental step and marked with $\times$.   } 
\label{figure_tsne}
\end{figure*} 
 
In Figure~\ref{figure_tsne}, we employ t-SNE~\cite{van2008visualizing} to visually analyze the feature representations on the CIFAR B0 Inc5 benchmark. Specifically, distinct marker symbols and colors are utilized to symbolize different class samples. Five classes from first task are marked with circles "$\circ$", and other five classes in the last incremental task are marked with "$\times$". Comparison with the baseline model, our model's experiential visualization demonstrates superior separability. Particularly, for the category represented by the red $\times$ markers, the visualization space in the baseline model exhibits noticeable dispersion, whereas our results predominantly cluster within a cohesive range.

\section{Conclusion and Discussion}

In this paper, we introduce a novel pre-trained models (PTMs) based continual class-incremental learning (CIL) method, termed FeTT model. Specifically, we adopt the fine-tune-then-frozen paradigm to perform parameter efficient fine-tuning (PEFT) in the first task to reduce the domain gap between the PTMs and downstream tasks. Subsequently, the model is froze and updates the prototypes for classification to mitigate mismatched training caused by class marginal distributions. In addition, feature transformation tuning is employed to adjust the scale of feature channels for further enhancing model performance without incurring any additional training or parameter costs. Extensive experimental results show that our model obtains superior performance on CL benchmarks. 

\noindent\textbf{Limitations and future work.}  
(1) This paper proposes to employ PTMs in the continual learning scenario with different tuning strategies. However, further exploration of the performance of different PTMs, such as multimodal vision-language models or ensemble model strategies, may represent a promising new direction for future research. (2) We observe significant advantages in directly learning class prototypes under the fine-tune-then-frozen paradigm for resisting catastrophic forgetting. However, this approach may limit the plasticity of the model, as it relies entirely on the discriminative space stability of excellent PTMs to achieve classification. The trade-off between the plasticity and stability of PTMs may be a potential new research direction. 

\bibliographystyle{unsrt}  
\bibliography{refs}  

\begin{thebibliography}{10}

\bibitem{DBLP:journals/corr/abs-1904-07734}
Gido~M. van~de Ven and Andreas~S. Tolias.
\newblock Three scenarios for continual learning.
\newblock {\em CoRR}, abs/1904.07734, 2019.

\bibitem{DBLP:conf/cvpr/MirzaMPB22}
Muhammad~Jehanzeb Mirza, Marc Masana, Horst Possegger, and Horst Bischof.
\newblock An efficient domain-incremental learning approach to drive in all weather conditions.
\newblock In {\em {IEEE/CVF} Conference on Computer Vision and Pattern Recognition Workshops, {CVPR} Workshops 2022}, pages 3000--3010. {IEEE}, 2022.

\bibitem{DBLP:journals/pami/LangeAMPJLST22}
Matthias~De Lange, Rahaf Aljundi, Marc Masana, Sarah Parisot, Xu~Jia, Ales Leonardis, Gregory~G. Slabaugh, and Tinne Tuytelaars.
\newblock A continual learning survey: Defying forgetting in classification tasks.
\newblock {\em {IEEE} Trans. Pattern Anal. Mach. Intell.}, 44(7):3366--3385, 2022.

\bibitem{DBLP:journals/corr/abs-2302-03648}
Da{-}Wei Zhou, Qi{-}Wei Wang, Zhi{-}Hong Qi, Han{-}Jia Ye, De{-}Chuan Zhan, and Ziwei Liu.
\newblock Deep class-incremental learning: {A} survey.
\newblock {\em CoRR}, abs/2302.03648, 2023.

\bibitem{MCCLOSKEY1989109}
Michael McCloskey and Neal~J. Cohen.
\newblock Catastrophic interference in connectionist networks: The sequential learning problem.
\newblock In {\em Psychology of Learning and Motivation}, volume~24, pages 109--165. Academic Press, 1989.

\bibitem{DBLP:journals/corr/KirkpatrickPRVD16}
James Kirkpatrick, Razvan Pascanu, Neil Rabinowitz, Joel Veness, Guillaume Desjardins, Andrei~A. Rusu, Kieran Milan, John Quan, Tiago Ramalho, Agnieszka Grabska-Barwinska, Demis Hassabis, Claudia Clopath, Dharshan Kumaran, and Raia Hadsell.
\newblock Overcoming catastrophic forgetting in neural networks.
\newblock {\em Proceedings of the National Academy of Sciences}, 114(13):3521--3526, 2017.

\bibitem{DBLP:conf/eccv/AljundiBERT18}
Rahaf Aljundi, Francesca Babiloni, Mohamed Elhoseiny, Marcus Rohrbach, and Tinne Tuytelaars.
\newblock Memory aware synapses: Learning what (not) to forget.
\newblock In {\em European Conference on Computer Vision, {ECCV} 2018}, volume 11207, pages 144--161. Springer, 2018.

\bibitem{DBLP:journals/corr/HintonVD15}
Geoffrey~E. Hinton, Oriol Vinyals, and Jeffrey Dean.
\newblock Distilling the knowledge in a neural network.
\newblock {\em CoRR}, abs/1503.02531, 2015.

\bibitem{DBLP:conf/eccv/LiH16}
Zhizhong Li and Derek Hoiem.
\newblock Learning without forgetting.
\newblock In {\em European Conference on Computer Vision, {ECCV} 2016}, volume 9908, pages 614--629. Springer, 2016.

\bibitem{DBLP:conf/cvpr/RebuffiKSL17}
Sylvestre{-}Alvise Rebuffi, Alexander Kolesnikov, Georg Sperl, and Christoph~H. Lampert.
\newblock icarl: Incremental classifier and representation learning.
\newblock In {\em 2017 {IEEE} Conference on Computer Vision and Pattern Recognition, {CVPR} 2017}, pages 5533--5542. {IEEE} Computer Society, 2017.

\bibitem{DBLP:conf/nips/Lopez-PazR17}
David Lopez{-}Paz and Marc'Aurelio Ranzato.
\newblock Gradient episodic memory for continual learning.
\newblock In {\em Advances in Neural Information Processing Systems 30, {NeurIPS} 2017}, pages 6467--6476, 2017.

\bibitem{DBLP:conf/aaai/QiangH0L0Z23}
Sunyuan Qiang, Jiayi Hou, Jun Wan, Yanyan Liang, Zhen Lei, and Du~Zhang.
\newblock Mixture uniform distribution modeling and asymmetric mix distillation for class incremental learning.
\newblock In {\em Thirty-Seventh {AAAI} Conference on Artificial Intelligence, {AAAI} 2023}, pages 9498--9506. {AAAI} Press, 2023.

\bibitem{DBLP:conf/cvpr/YanX021}
Shipeng Yan, Jiangwei Xie, and Xuming He.
\newblock {DER:} dynamically expandable representation for class incremental learning.
\newblock In {\em {IEEE} Conference on Computer Vision and Pattern Recognition, {CVPR} 2021}, pages 3014--3023. Computer Vision Foundation / {IEEE}, 2021.

\bibitem{DBLP:conf/eccv/WangZYZ22}
Fu{-}Yun Wang, Da{-}Wei Zhou, Han{-}Jia Ye, and De{-}Chuan Zhan.
\newblock {FOSTER:} feature boosting and compression for class-incremental learning.
\newblock In {\em uropean Conference on Computer Vision, {ECCV} 2022}, volume 13685, pages 398--414. Springer, 2022.

\bibitem{qiang2024dynamic}
Sunyuan Qiang, Yanyan Liang, Jun Wan, and Du~Zhang.
\newblock Dynamic feature learning and matching for class-incremental learning.
\newblock {\em CoRR}, abs/2405.08533, 2024.

\bibitem{DBLP:conf/cvpr/0002ZL0SRSPDP22}
Zifeng Wang, Zizhao Zhang, Chen{-}Yu Lee, Han Zhang, Ruoxi Sun, Xiaoqi Ren, Guolong Su, Vincent Perot, Jennifer~G. Dy, and Tomas Pfister.
\newblock Learning to prompt for continual learning.
\newblock In {\em {IEEE/CVF} Conference on Computer Vision and Pattern Recognition, {CVPR} 2022}, pages 139--149. {IEEE}, 2022.

\bibitem{DBLP:conf/eccv/0002ZESZLRSPDP22}
Zifeng Wang, Zizhao Zhang, Sayna Ebrahimi, Ruoxi Sun, Han Zhang, Chen{-}Yu Lee, Xiaoqi Ren, Guolong Su, Vincent Perot, Jennifer~G. Dy, and Tomas Pfister.
\newblock Dualprompt: Complementary prompting for rehearsal-free continual learning.
\newblock In {\em European Conference on Computer Vision, {ECCV} 2022}, volume 13686, pages 631--648. Springer, 2022.

\bibitem{DBLP:conf/cvpr/SmithKGCKAPFK23}
James~Seale Smith, Leonid Karlinsky, Vyshnavi Gutta, Paola Cascante{-}Bonilla, Donghyun Kim, Assaf Arbelle, Rameswar Panda, Rog{\'{e}}rio Feris, and Zsolt Kira.
\newblock Coda-prompt: Continual decomposed attention-based prompting for rehearsal-free continual learning.
\newblock In {\em {IEEE/CVF} Conference on Computer Vision and Pattern Recognition, {CVPR} 2023}, pages 11909--11919. {IEEE}, 2023.

\bibitem{DBLP:journals/corr/abs-2303-07338}
Da{-}Wei Zhou, Han{-}Jia Ye, De{-}Chuan Zhan, and Ziwei Liu.
\newblock Revisiting class-incremental learning with pre-trained models: Generalizability and adaptivity are all you need.
\newblock {\em CoRR}, abs/2303.07338, 2023.

\bibitem{DBLP:journals/nn/BelouadahPK21}
Eden Belouadah, Adrian Popescu, and Ioannis Kanellos.
\newblock A comprehensive study of class incremental learning algorithms for visual tasks.
\newblock {\em Neural Networks}, 135:38--54, 2021.

\bibitem{DBLP:journals/corr/abs-2401-16386}
Da{-}Wei Zhou, Hai{-}Long Sun, Jingyi Ning, Han{-}Jia Ye, and De{-}Chuan Zhan.
\newblock Continual learning with pre-trained models: {A} survey.
\newblock {\em CoRR}, abs/2401.16386, 2024.

\bibitem{DBLP:journals/pr/LiWSX23}
Xiaorong Li, Shipeng Wang, Jian Sun, and Zongben Xu.
\newblock Memory efficient data-free distillation for continual learning.
\newblock {\em Pattern Recognit.}, 144:109875, 2023.

\bibitem{WU2024110440}
Ran Wu, Huanyu Liu, Zongcheng Yue, Jun-Bao Li, and Chiu-Wing Sham.
\newblock Hyper-feature aggregation and relaxed distillation for class incremental learning.
\newblock {\em Pattern Recognit.}, 152:110440, 2024.

\bibitem{SONG2024110506}
Jialun Song, Jian Chen, and Lan Du.
\newblock Rebalancing network with knowledge stability for class incremental learning.
\newblock {\em Pattern Recognit.}, 153:110506, 2024.

\bibitem{DBLP:conf/nips/ShinLKK17}
Hanul Shin, Jung~Kwon Lee, Jaehong Kim, and Jiwon Kim.
\newblock Continual learning with deep generative replay.
\newblock In {\em Advances in Neural Information Processing Systems 30, {NeurIPS}}, pages 2990--2999, 2017.

\bibitem{DBLP:conf/icml/GaoL23a}
Rui Gao and Weiwei Liu.
\newblock {DDGR:} continual learning with deep diffusion-based generative replay.
\newblock In {\em International Conference on Machine Learning, {ICML} 2023}, volume 202, pages 10744--10763. {PMLR}, 2023.

\bibitem{DBLP:journals/pr/LaoMTDFH21}
Qicheng Lao, Mehrzad Mortazavi, Marzieh Tahaei, Francis Dutil, Thomas Fevens, and Mohammad Havaei.
\newblock Focl: Feature-oriented continual learning for generative models.
\newblock {\em Pattern Recognit.}, 120:108127, 2021.

\bibitem{DBLP:conf/eccv/DouillardCORV20}
Arthur Douillard, Matthieu Cord, Charles Ollion, Thomas Robert, and Eduardo Valle.
\newblock Podnet: Pooled outputs distillation for small-tasks incremental learning.
\newblock In {\em European Conference on Computer Vision, {ECCV} 2020}, volume 12365, pages 86--102. Springer, 2020.

\bibitem{DBLP:conf/iclr/DosovitskiyB0WZ21}
Alexey Dosovitskiy, Lucas Beyer, Alexander Kolesnikov, Dirk Weissenborn, Xiaohua Zhai, Thomas Unterthiner, Mostafa Dehghani, Matthias Minderer, Georg Heigold, Sylvain Gelly, Jakob Uszkoreit, and Neil Houlsby.
\newblock An image is worth 16x16 words: Transformers for image recognition at scale.
\newblock In {\em 9th International Conference on Learning Representations, {ICLR} 2021}. OpenReview.net, 2021.

\bibitem{DBLP:conf/icml/RadfordKHRGASAM21}
Alec Radford, Jong~Wook Kim, Chris Hallacy, Aditya Ramesh, Gabriel Goh, Sandhini Agarwal, Girish Sastry, Amanda Askell, Pamela Mishkin, Jack Clark, Gretchen Krueger, and Ilya Sutskever.
\newblock Learning transferable visual models from natural language supervision.
\newblock In {\em Proceedings of the 38th International Conference on Machine Learning, {ICML} 2021}, volume 139, pages 8748--8763. {PMLR}, 2021.

\bibitem{DBLP:conf/iccv/ZhengMWQYY23}
Zangwei Zheng, Mingyuan Ma, Kai Wang, Ziheng Qin, Xiangyu Yue, and Yang You.
\newblock Preventing zero-shot transfer degradation in continual learning of vision-language models.
\newblock In {\em {IEEE/CVF} International Conference on Computer Vision, {ICCV} 2023}, pages 19068--19079. {IEEE}, 2023.

\bibitem{DBLP:journals/corr/abs-2403-11549}
Jiazuo Yu, Yunzhi Zhuge, Lu~Zhang, Ping Hu, Dong Wang, Huchuan Lu, and You He.
\newblock Boosting continual learning of vision-language models via mixture-of-experts adapters.
\newblock {\em CoRR}, abs/2403.11549, 2024.

\bibitem{DBLP:conf/iccv/ZhangWKCW23}
Gengwei Zhang, Liyuan Wang, Guoliang Kang, Ling Chen, and Yunchao Wei.
\newblock {SLCA:} slow learner with classifier alignment for continual learning on a pre-trained model.
\newblock In {\em {IEEE/CVF} International Conference on Computer Vision, {ICCV} 2023}, pages 19091--19101. {IEEE}, 2023.

\bibitem{DBLP:conf/nips/McDonnellGPAH23}
Mark~D. McDonnell, Dong Gong, Amin Parvaneh, Ehsan Abbasnejad, and Anton van~den Hengel.
\newblock Ranpac: Random projections and pre-trained models for continual learning.
\newblock In {\em Advances in Neural Information Processing Systems 36, {NeurIPS} 2023}, 2023.

\bibitem{DBLP:conf/naacl/DevlinCLT19}
Jacob Devlin, Ming{-}Wei Chang, Kenton Lee, and Kristina Toutanova.
\newblock {BERT:} pre-training of deep bidirectional transformers for language understanding.
\newblock In {\em Proceedings of the 2019 Conference of the North American Chapter of the Association for Computational Linguistics: Human Language Technologies, {NAACL-HLT} 2019}, pages 4171--4186. Association for Computational Linguistics, 2019.

\bibitem{radford2019language}
Alec Radford, Jeffrey Wu, Rewon Child, David Luan, Dario Amodei, Ilya Sutskever, et~al.
\newblock Language models are unsupervised multitask learners.
\newblock {\em OpenAI blog}, 1(8):9, 2019.

\bibitem{DBLP:conf/icml/ChenK0H20}
Ting Chen, Simon Kornblith, Mohammad Norouzi, and Geoffrey~E. Hinton.
\newblock A simple framework for contrastive learning of visual representations.
\newblock In {\em Proceedings of the 37th International Conference on Machine Learning, {ICML} 2020}, volume 119, pages 1597--1607. {PMLR}, 2020.

\bibitem{DBLP:conf/iccv/HeGD19}
Kaiming He, Ross~B. Girshick, and Piotr Doll{\'{a}}r.
\newblock Rethinking imagenet pre-training.
\newblock In {\em 2019 {IEEE/CVF} International Conference on Computer Vision, {ICCV} 2019}, pages 4917--4926. {IEEE}, 2019.

\bibitem{DBLP:conf/cvpr/He0WXG20}
Kaiming He, Haoqi Fan, Yuxin Wu, Saining Xie, and Ross~B. Girshick.
\newblock Momentum contrast for unsupervised visual representation learning.
\newblock In {\em 2020 {IEEE/CVF} Conference on Computer Vision and Pattern Recognition, {CVPR} 2020}, pages 9726--9735. Computer Vision Foundation / {IEEE}, 2020.

\bibitem{DBLP:journals/corr/abs-2312-12148}
Lingling Xu, Haoran Xie, Si{-}Zhao~Joe Qin, Xiaohui Tao, and Fu~Lee Wang.
\newblock Parameter-efficient fine-tuning methods for pretrained language models: {A} critical review and assessment.
\newblock {\em CoRR}, abs/2312.12148, 2023.

\bibitem{DBLP:conf/eccv/JiaTCCBHL22}
Menglin Jia, Luming Tang, Bor{-}Chun Chen, Claire Cardie, Serge~J. Belongie, Bharath Hariharan, and Ser{-}Nam Lim.
\newblock Visual prompt tuning.
\newblock In {\em European Conference on Computer Vision, {ECCV} 2022}, volume 13693, pages 709--727. Springer, 2022.

\bibitem{DBLP:conf/iclr/HuSWALWWC22}
Edward~J. Hu, Yelong Shen, Phillip Wallis, Zeyuan Allen{-}Zhu, Yuanzhi Li, Shean Wang, Lu~Wang, and Weizhu Chen.
\newblock Lora: Low-rank adaptation of large language models.
\newblock In {\em The Tenth International Conference on Learning Representations, {ICLR} 2022}. OpenReview.net, 2022.

\bibitem{DBLP:conf/nips/LianZFW22}
Dongze Lian, Daquan Zhou, Jiashi Feng, and Xinchao Wang.
\newblock Scaling {\&} shifting your features: {A} new baseline for efficient model tuning.
\newblock In {\em Advances in Neural Information Processing Systems 35, {NeurIPS} 2022}, 2022.

\bibitem{DBLP:conf/nips/ChenGTWSWL22}
Shoufa Chen, Chongjian Ge, Zhan Tong, Jiangliu Wang, Yibing Song, Jue Wang, and Ping Luo.
\newblock Adaptformer: Adapting vision transformers for scalable visual recognition.
\newblock In {\em Advances in Neural Information Processing Systems 35, {NeurIPS} 2022}, 2022.

\bibitem{DBLP:conf/nips/VaswaniSPUJGKP17}
Ashish Vaswani, Noam Shazeer, Niki Parmar, Jakob Uszkoreit, Llion Jones, Aidan~N. Gomez, Lukasz Kaiser, and Illia Polosukhin.
\newblock Attention is all you need.
\newblock In {\em Advances in Neural Information Processing Systems 30, {NeurIPS} 2017}, pages 5998--6008, 2017.

\bibitem{DBLP:conf/icml/IoffeS15}
Sergey Ioffe and Christian Szegedy.
\newblock Batch normalization: Accelerating deep network training by reducing internal covariate shift.
\newblock In {\em Proceedings of the 32nd International Conference on Machine Learning, {ICML} 2015}, volume~37, pages 448--456. JMLR.org, 2015.

\bibitem{DBLP:conf/iclr/YangLX21}
Shuo Yang, Lu~Liu, and Min Xu.
\newblock Free lunch for few-shot learning: Distribution calibration.
\newblock In {\em 9th International Conference on Learning Representations, {ICLR} 2021}. OpenReview.net, 2021.

\bibitem{DBLP:conf/icml/LuoXX22}
Xu~Luo, Jing Xu, and Zenglin Xu.
\newblock Channel importance matters in few-shot image classification.
\newblock In {\em International Conference on Machine Learning, {ICML} 2022}, volume 162, pages 14542--14559. {PMLR}, 2022.

\bibitem{DBLP:journals/corr/abs-2309-11497}
Chenyang Si, Ziqi Huang, Yuming Jiang, and Ziwei Liu.
\newblock Freeu: Free lunch in diffusion u-net.
\newblock {\em CoRR}, abs/2309.11497, 2023.

\bibitem{tukey1977exploratory}
John~Wilder Tukey et~al.
\newblock {\em Exploratory data analysis}, volume~2.
\newblock Springer, 1977.

\bibitem{DBLP:conf/cvpr/ZhouKLOT16}
Bolei Zhou, Aditya Khosla, {\`{A}}gata Lapedriza, Aude Oliva, and Antonio Torralba.
\newblock Learning deep features for discriminative localization.
\newblock In {\em 2016 {IEEE} Conference on Computer Vision and Pattern Recognition, {CVPR} 2016}, pages 2921--2929. {IEEE} Computer Society, 2016.

\bibitem{DBLP:conf/iclr/BaiZJXM021}
Yang Bai, Yuyuan Zeng, Yong Jiang, Shu{-}Tao Xia, Xingjun Ma, and Yisen Wang.
\newblock Improving adversarial robustness via channel-wise activation suppressing.
\newblock In {\em 9th International Conference on Learning Representations, {ICLR} 2021}. OpenReview.net, 2021.

\bibitem{DBLP:conf/nips/Vapnik91}
Vladimir Vapnik.
\newblock Principles of risk minimization for learning theory.
\newblock In {\em Advances in Neural Information Processing Systems 4, {[NeuIPS} 1991}, pages 831--838. Morgan Kaufmann, 1991.

\bibitem{DBLP:conf/cvpr/0004TLHWCJ020}
Lu~Yu, Bartlomiej Twardowski, Xialei Liu, Luis Herranz, Kai Wang, Yongmei Cheng, Shangling Jui, and Joost van~de Weijer.
\newblock Semantic drift compensation for class-incremental learning.
\newblock In {\em 2020 {IEEE/CVF} Conference on Computer Vision and Pattern Recognition, {CVPR} 2020}, pages 6980--6989. Computer Vision Foundation / {IEEE}, 2020.

\bibitem{krizhevsky2009learning}
Alex Krizhevsky, Geoffrey Hinton, et~al.
\newblock Learning multiple layers of features from tiny images.
\newblock {\em Toronto, ON, Canada}, 2009.

\bibitem{wah2011caltech}
Catherine Wah, Steve Branson, Peter Welinder, Pietro Perona, and Serge Belongie.
\newblock The caltech-ucsd birds-200-2011 dataset.
\newblock {\em California Institute of Technology}, 2011.

\bibitem{DBLP:conf/cvpr/HendrycksZBSS21}
Dan Hendrycks, Kevin Zhao, Steven Basart, Jacob Steinhardt, and Dawn Song.
\newblock Natural adversarial examples.
\newblock In {\em {IEEE} Conference on Computer Vision and Pattern Recognition, {CVPR} 2021}, pages 15262--15271. Computer Vision Foundation / {IEEE}, 2021.

\bibitem{DBLP:conf/iccv/HendrycksBMKWDD21}
Dan Hendrycks, Steven Basart, Norman Mu, Saurav Kadavath, Frank Wang, Evan Dorundo, Rahul Desai, Tyler Zhu, Samyak Parajuli, Mike Guo, Dawn Song, Jacob Steinhardt, and Justin Gilmer.
\newblock The many faces of robustness: {A} critical analysis of out-of-distribution generalization.
\newblock In {\em 2021 {IEEE/CVF} International Conference on Computer Vision, {ICCV} 2021}, pages 8320--8329. {IEEE}, 2021.

\bibitem{DBLP:conf/nips/BarbuMALWGTK19}
Andrei Barbu, David Mayo, Julian Alverio, William Luo, Christopher Wang, Dan Gutfreund, Josh Tenenbaum, and Boris Katz.
\newblock Objectnet: {A} large-scale bias-controlled dataset for pushing the limits of object recognition models.
\newblock In {\em Advances in Neural Information Processing Systems, {NeurIPS} 2019}, pages 9448--9458, 2019.

\bibitem{zhai2019large}
Xiaohua Zhai, Joan Puigcerver, Alexander Kolesnikov, Pierre Ruyssen, Carlos Riquelme, Mario Lucic, Josip Djolonga, Andre~Susano Pinto, Maxim Neumann, Alexey Dosovitskiy, et~al.
\newblock A large-scale study of representation learning with the visual task adaptation benchmark.
\newblock {\em arXiv preprint arXiv:1910.04867}, 2019.

\bibitem{DBLP:conf/cvpr/DengDSLL009}
Jia Deng, Wei Dong, Richard Socher, Li{-}Jia Li, Kai Li, and Li~Fei{-}Fei.
\newblock Imagenet: {A} large-scale hierarchical image database.
\newblock In {\em 2009 {IEEE} Computer Society Conference on Computer Vision and Pattern Recognition {(CVPR} 2009)}, pages 248--255. {IEEE} Computer Society, 2009.

\bibitem{DBLP:conf/nips/PaszkeGMLBCKLGA19}
Adam Paszke, Sam Gross, Francisco Massa, Adam Lerer, James Bradbury, Gregory Chanan, Trevor Killeen, Zeming Lin, Natalia Gimelshein, Luca Antiga, Alban Desmaison, Andreas K{\"{o}}pf, Edward~Z. Yang, Zachary DeVito, Martin Raison, Alykhan Tejani, Sasank Chilamkurthy, Benoit Steiner, Lu~Fang, Junjie Bai, and Soumith Chintala.
\newblock Pytorch: An imperative style, high-performance deep learning library.
\newblock In {\em Advances in Neural Information Processing Systems 32, {NeurIPS} 2019}, pages 8024--8035, 2019.

\bibitem{DBLP:conf/nips/RidnikBNZ21}
Tal Ridnik, Emanuel~Ben Baruch, Asaf Noy, and Lihi Zelnik.
\newblock Imagenet-21k pretraining for the masses.
\newblock In {\em Proceedings of the Neural Information Processing Systems Track on Datasets and Benchmarks, NeurIPS Datasets and Benchmarks 2021}, 2021.

\bibitem{van2008visualizing}
Laurens Van~der Maaten and Geoffrey Hinton.
\newblock Visualizing data using t-sne.
\newblock {\em J. Mach. Learn. Res.}, 9(86):2579--2605, 2008.

\end{thebibliography}

\end{document}